%% file: paper.tex
\pdfoutput=1
\documentclass{article}
\usepackage[final,nonatbib]{neurips_2022}
\input{common}

\title{%
    \attack: Improving Ensemble Robustness Evaluation
    with Model-Reweighing Attack}

\author{%
    Yunrui Yu\thanks{%
        Xitong Gao and Yunrui Yu
        contributed equally to this work.
    } \\
    University of Macau \\
    {\small Macau SAR, China} \\
    \email{yb97445@um.edu.mo}
    \And
    Xitong Gao\footnotemark[1] \\
    \hspace{-7pt}
    Shenzhen Institute of Advanced Technology,
    \hspace{-7pt} \\
    {\small Chinese Academy of Sciences,
    Shenzhen, China} \\
    \email{xt.gao@siat.ac.cn} \\
    \And
    Cheng-Zhong Xu\thanks{%
        Correspondence to:
        Cheng-Zhong Xu (\email{cz.xu@um.edu.mo}).
    } \\
    University of Macau \\
    {\small Macau SAR, China} \\
    \email{czxu@um.edu.mo}
}

\begin{document}
    \maketitle
    \setcounter{footnote}{0}
    \input{abstract}
    \input{intro}
    \input{related}
    \input{method}
    \input{results}

    \input{conclusion}
    {
        \small
        \bibliographystyle{plain}
        \bibliography{references}
    }
    \clearpage
    \appendix
    \input{appendix}
\end{document}

%% file: common.tex
\usepackage[utf8]{inputenc} 
\usepackage[T1]{fontenc}    
\usepackage{hyperref}       
\usepackage{url}            
\usepackage{booktabs}       
\usepackage{amsfonts}       
\usepackage{nicefrac}       
\usepackage{microtype}      
\usepackage{xcolor}         

\usepackage{caption,subcaption}
\usepackage{placeins}
\usepackage{times,xspace}
\usepackage{graphicx,grffile}
\usepackage{amsmath,amssymb,bm,mathtools,yhmath}
\usepackage{algorithm,algpseudocode}
\usepackage{adjustbox,multirow,siunitx}
\usepackage{cleveref}
\usepackage{pgfplots,pgfplotstable}

\usepgfplotslibrary{colorbrewer,external}
\pgfplotsset{compat=newest}
\graphicspath{{figures/}}

\makeatletter
\DeclareRobustCommand\onedot{\futurelet\@let@token\@onedot}
\def\@onedot{\ifx\@let@token.\else.\null\fi\xspace}
\newcommand{\eg}{\emph{e.g\@\onedot}}
\newcommand{\etc}{\emph{etc\@\onedot}}

\newcommand{\ie}{\emph{i.e\@\onedot}}
\newcommand{\vs}{\texorpdfstring{\emph{vs\@\onedot}}{vs.}}

\newcommand{\wrt}{\emph{w.r.t\@\onedot}}

\newcommand{\numero}[1]{No\@.\ {#1}}

\newcommand{\supdagger}{\textsuperscript{\textdagger}}

\makeatletter%
\newcolumntype{L}{D{.}{.}{2,2}}
\newcolumntype{B}[3]{>{\boldmath\DC@{#1}{#2}{#3}}c<{\DC@end}}
\newcommand{\tbox}[1]{\begin{tabular}[c]{@{}c@{}}#1\end{tabular}}

\newcommand{\tcenter}[1]{\multicolumn{1}{c}{{#1}}}
\newcommand{\thead}[1]{\multicolumn{1}{c}{\textbf{#1}}}
\newcommand{\tna}{\tcenter{---}}
\newcommand{\bfzero}{\textbf{0.00}}
\makeatother%

\newcommand{\ordinal}[1]{{#1}\textsuperscript{th}}

\newcommand{\cifarx}{CIFAR-10}
\newcommand{\cifarc}{CIFAR-100}

\newcommand\attack{{MORA}}
\newcommand\attackmt{{MORA}\textsuperscript{mt}}
\newcommand{\cw}{C\&W}
\newcommand{\aaa}{\( \mathrm{A}^3 \)}

\DeclarePairedDelimiter{\parens}{\lparen}{\rparen}
\DeclarePairedDelimiter{\angles}{\langle}{\rangle}
\DeclarePairedDelimiter{\bracks}{[}{]}
\DeclarePairedDelimiter{\braces}{\{}{\}}

\DeclarePairedDelimiter{\norm}{\lVert}{\rVert}
\newcommand{\detach}{\mathrm{detach}}
\newcommand{\ssum}{\textstyle\sum}
\newcommand{\shat}[1]{\vphantom{#1}\smash[t]{\hat{#1}}}
\newcommand{\linf}{\ensuremath\ell^\infty}

\newcommand{\expect}[2][]{\mathsf{E}_{#1}\left[{#2}\right]}

\newcommand{\uniform}[1]{\mathcal{U}\left({#1}\right)}
\newcommand{\realset}{\mathbb{R}}
\newcommand{\idx}[2]{{#1}^{\scriptscriptstyle[{#2}]}}
\newcommand{\x}{\mathbf{x}}

\newcommand{\advset}{\mathcal{A}_{\epsilon, \x}}
\newcommand{\xadv}{\hat\x}
\newcommand{\y}{y}
\newcommand{\hy}{{\shat{y}}}
\newcommand{\z}{\mathbf{z}}
\newcommand{\g}{\bm{g}}
\newcommand{\f}[1]{{f}_{#1}}
\newcommand{\E}{\mathrm{E}}
\newcommand{\m}{\bm{\mu}}
\newcommand{\logit}[1]{\idx{\z}{#1}}

\newcommand{\dl}[1]{\idx{k}{#1}}
\newcommand{\imp}[1]{\idx{\lambda}{#1}_\tau}

\newcommand{\allw}{{\bm\theta}}

\newcommand{\inputset}{\mathcal{I}}
\newcommand{\classset}{\mathcal{C}}
\newcommand{\intensorset}{%
    \ensuremath{\left[0, 1\right]}^{{C}\times{H}\times{W}}}
\newcommand{\outputset}{\realset^K}
\newcommand{\stepsize}{\alpha}
\newcommand{\momentum}{\nu}

\DeclareMathOperator{\scenorm}{\mathrm{norm}}
\DeclareMathOperator{\softmax}{\mathrm{softmax}}
\DeclareMathOperator{\wta}{\mathrm{wta}}
\DeclareMathOperator{\softwta}{\mathrm{softwta}}
\DeclareMathOperator{\id}{\mathrm{id}}
\DeclareMathOperator{\ensop}{\mathrm{ens}}
\DeclareMathOperator{\loss}{\mathcal{L}}
\DeclareMathOperator{\attackloss}{%
    \mathcal{L}^\mathrm{mora}_{\beta, \tau}}
\DeclareMathOperator{\sceloss}{%
    \mathcal{L}^\mathrm{sce}}

\DeclareMathOperator{\project}{%
    \mathcal{P}_{\epsilon, \x}}
\DeclareMathOperator{\sign}{\mathrm{sign}}

\DeclareMathOperator{\indicator}{{1}}

\everydisplay{\textstyle}

\newcommand{\subplot}[3][]{%
    \begin{subfigure}[b]{0.32\textwidth}
        \centering\includegraphics[
            scale=0.53, trim=10pt 5pt 10pt 5pt, clip
        ]{#2}\vspace{-5pt}%
        \caption{#3.}\label{fig:#1#2}
    \end{subfigure}%
}

\newcommand{\email}[1]{\href{mailto:#1}{\texttt{#1}}}

%% file: abstract.tex
\begin{abstract}
    Adversarial attacks
    can deceive neural networks
    by adding tiny perturbations to their input data.
    Ensemble defenses,
    which are trained
    to minimize attack transferability among sub-models,
    offer a promising research direction
    to improve robustness against such attacks
    while maintaining a high accuracy on natural inputs.
    We discover, however,
    that recent state-of-the-art (SOTA)
    adversarial attack strategies
    cannot reliably evaluate ensemble defenses,
    sizeably overestimating their robustness.
    This paper identifies the two factors
    that contribute to this behavior.
    First,
    these defenses form ensembles
    that are notably difficult
    for existing gradient-based method
    to attack,
    due to gradient obfuscation.
    Second,
    ensemble defenses diversify sub-model gradients,
    presenting a challenge
    to defeat all sub-models simultaneously,
    simply summing their contributions
    may counteract the overall attack objective;
    yet, we observe that ensemble may still be fooled
    despite most sub-models being correct.
    We therefore introduce \attack,
    a model-reweighing attack
    to steer adversarial example synthesis
    by reweighing the importance of sub-model gradients.
    \attack{} finds that recent ensemble defenses
    all exhibit varying degrees of overestimated robustness.
    Comparing it against recent SOTA white-box attacks,
    it can converge orders of magnitude faster
    while achieving higher attack success rates
    across all ensemble models examined
    with three different ensemble modes
    (\ie{}, ensembling by either softmax, voting or logits).
    In particular,
    most ensemble defenses
    exhibit near or exactly \( 0\% \) robustness
    against \attack{}
    with \( \linf \) perturbation
    within \( 0.02 \) on \cifarx,
    and \( 0.01 \) on \cifarc.
    We make \attack{} open source
    with reproducible results and pre-trained models;
    and provide a leaderboard
    of ensemble defenses
    under various attack strategies\footnote{%
    \url{https://github.com/lafeat/mora}.}.
\end{abstract}

%% file: intro.tex
\section{Introduction}\label{sec:intro}

Many safety-critical applications,
such as autonomous robots~\cite{zhu2021can},
self-driving~\cite{eykholt2018robust},
search engines~\cite{tolias2019targeted}, \etc{}
are becoming increasingly powerful and reliant
on deep neural networks (DNNs).
Despite the monumental success of DNNs
on these applications,
they are highly susceptible
to adversarial examples ---
an attacker can add tiny delibrate perturbations
to the input data,
misleading the model
into giving incorrect results~\cite{szegedy14,goodfellow15}.
Such adversarial attacks
could pose a significant threat
to the safety and reliability
of deep learning applications.

To mitigate this threat,
many defense strategies~\cite{madry18,zhang19trades,carmon19}
based on adversarial training~\cite{madry18}
have been proposed
to improve model robustness.
Adversarial training, however,
gains robustness at the expense of model accuracy
on clean natural images~\cite{tsipras2018robustness}.
Ensemble defenses~\cite{
    pang2019adp,kariyappa2019gal,yang2020dverge,yang2021trs}
have thus emerged
to combine multiple predictions
from independent sub-models.
The intuition is
that an ensemble of models
can often lead to higher accuracy,
while learning to stop
adversarial example transfer among sub-models
may improve robustness against adversarial attacks.
This approach could potentially
offer a promising research direction
to improve model robustness
while preserving high accuracy on natural inputs.

\input{figures/motivation}
Yet surprisingly,
under the white-box threat model,
existing state-of-the-art (SOTA)
adversarial attacks
with strong performance
on conventional DNN models
performed poorly on ensemble models,
sizeably overestimating their robustness
(\Cref{fig:scatter/gal,fig:scatter/dverge}).
This also suggests, to some extent,
that ensemble defenses
may rely on two potential design flaws below
that cause obfuscated gradients~\cite{athalye2018obfuscated},
\ie{}, they are either deliberately non-differentiable,
or give no useful gradients,
thus inducing overestimated robustness:

(a) \emph{%
    Gradient obfuscation
    via ensemble-forming strategy.}
They typically form ensembles
by averaging probability vectors (softmax)
of sub-models,
and softmax operations
can easily cause gradient obfuscation.
While the model's actual robustness
is pertinent to the strategy
used to form an ensemble,
this indicates
that gradient-based attacks
have to \emph{also} leverage this effectively.

(b) \emph{Gradient diversification.}
Motivated by the reasoning
that a majority of sub-models
may need to be fooled
for successful attacks,
they learn to reduce adversarial transferability
among sub-models,
often via gradient diversification.
This intuitively
causes sub-models to counteract each other,
averaging to a small or inaccurate overall gradient.
Attacking only the ensemble loss
would fool most sub-models,
but the ensemble may remain still correct;
conversely, it is actually possible to fool an ensemble,
despite the majority of its sub-models
giving correct predictions (\Cref{fig:num_models/logits}).

From the above observations,
it is perceivable that
the practical evaluation of ensemble robustness
cannot be solely done
by treating such models holistically.
To this end,
this paper introduces \attack,
\underline{mo}del-\underline{r}eweighing \underline{a}ttack,
to adaptively adjust the importance
of sub-models in attack iterations.
Sub-models are reweighed
according to their respective ``ease of attack'',
which is in turn evaluated
by the gradient of the difference
of ensemble classification outputs
\wrt{} the ones of individual sub-models.
Pushing the limits of the current SOTA
in ensemble robustness evaluation,
it draws inspiration
from recent effective attack tactics,
\eg, momentum~\cite{dong18momentum,croce20aa},
step size schedule~\cite{croce20aa,ye2022aaa},
loss normalization~\cite{lafeat},
and multiple targets~\cite{croce20aa,tramer2020adaptive}.
We summarize our contributions:
\begin{itemize}
    \item This paper presents the first extensive study
    on the robustness of ensemble defenses
    under multiple ensemble-forming strategies.

    \item By reweighing the importance weights of sub-models
    to steer adversarial example synthesis,
    we show that
    gradient-based attacks on ensemble defenses
    can often be orders of magnitude faster,
    while enjoying a higher success rate.


    \item
    Empirical results
    on a wide variety
    of different ensemble defenses
    show that \attack{}
    outperforms competing attacks
    in both performance and convergence rate.
    Finally, this paper
    provides extensive ablation of its components
    and sensitivity analyses of hyperparameters.
\end{itemize}

To our best knowledge,
\attack{}
is currently the strongest attack
against a wide range of ensemble defenses.
We make \attack{} open source
with reproducible results and pre-trained models;
moreover,
we maintain a leaderboard of ensemble defenses
under various attack strategies.

%% file: figures/motivation.tex
\begin{figure}[t]
    \centering%
    \subplot{scatter/gal}{GAL}
    \subplot{scatter/dverge}{Dverge}
    \subplot{num_models/logits}{Robust sub-models \vs{} attacks}
    \caption{%
        (\subref{fig:scatter/gal},\subref{fig:scatter/dverge})
        Existing attacks~\cite{
            madry18,carlini17,croce20aa,mao2021caa,ye2022aaa}
        with strong baselines
        are neither efficient
        in the number of model forward/backward passes,
        nor reliable in the estimation of ensemble robustness
        when compared with \attack.
        GAL~\cite{kariyappa2019gal}
        and Dverge~\cite{yang2020dverge}
        defenses are trained on \cifarx{}
        with 8 sub-models.
        We used \( \linf \) attacks
        within \( \epsilon = 0.01 \),
        ``Nominal'' is self-reported.
        (\subref{fig:num_models/logits})
        \attack{} can successfully fool
        logit-based ensembles
        (ADP~\cite{pang2019adp}, GAL and Dverge)
        even with the majority of their sub-models
        giving correct outputs
        (``\( A \rightarrow B \)''
         means using \( A \) to attack \( B \)
         for up to 100 iterations).
    }\label{fig:intro:motivation}
\end{figure}%

%% file: related.tex
\section{Preliminaries \& Related Work}

\subsection{Adversarial Attacks}

An adversarial example
adds a small perturbation,
typically bounded a small value
with \( \ell^p \) norms,
to the original image,
such that the model under attack
can be deceived into giving incorrect results.
The advent of adversarial attacks~\cite{szegedy14}
has piqued the interest of deep learning practitioners,
and revealed security concerns
of deep learning~\cite{%
    tramer2019adversarial,rosenberg2021adversarial},
improved GAN training~\cite{bashkirova2019adversarial}
transfer learning~\cite{utrera2021adversariallytrained,deng2021adversarial},
and DNN interpretability~\cite{ross2018improving},
\etc{}
Formally,
assuming a defending classifier
\( f: \inputset \to \outputset \),
taking an input image
\( \x \in \inputset = \intensorset \)
as input,
and \( \arg\max f\parens{\x} \)
predicts the correct class label \( y \in \classset \),
then an attacker attempts to find an adversarial example
\( \xadv \) in the set:
\begin{equation}
    \braces{
        \xadv \in \advset
        \colon
        \arg\max f\parens{\xadv} \neq y
    }.
    \label{eq:adversarial_example}
\end{equation}
Here,
\( \xadv \in \advset \)
constrains the adversarial example \( \xadv \)
to be within both the input space \( \inputset \)
and a small \( \epsilon \)-ball
of \( \ell^p \)-distance
from the original image \( \x \),
or equivalently \( \norm{\x - \xadv}_p \leq \epsilon \).
Satisfying the condition
\( \arg\max f\parens{\x} \neq y \)
means that \( f\parens{\x} \) fails
to give the expected correct classification \( y \).
We focus on the \( \ell^\infty \)
white-box threat model
commonly considered by the defenses
examined in this paper,
which grants the attacker completely access
to the internals of the defender,
including, for instance, its model architecture,
parameters, training algorithms, \etc{}

One of the popular and effective white-box attacks
used by many defenders
to evaluate their robustness
is \emph{projected gradient descent} (PGD)~\cite{madry18},
which finds adversarial examples
by maximizing the classification loss
with gradient descent:
\begin{equation}
    \xadv_{i + 1} = \project\parens{
        \xadv_i + \alpha_i \sign\parens{
            \nabla \loss\parens{f\parens{\xadv_i}, y}
        }
    },
\end{equation}
where
\( \loss \) is typically
the softmax cross-entropy (SCE) loss
used to train the model,
 \( \alpha_i \) is the step size,
and we let the initial
\( \xadv_0 \triangleq \project\parens{\x + \m} \).
The projection function \( \project\parens{\mathbf{v}} \)
constrains its input \( \mathbf{v} \)
to be within the feasible region \( \advset \),
and finally
\( \m \sim \uniform{-\epsilon, \epsilon} \)
is a uniformly distributed noise
bounded by \( \bracks{-\epsilon, \epsilon} \).
Besides PGD,
C\&W~\cite{carlini17}
is also a gradient-based attack
which, instead of projection,
indirectly constrains the search space
by regularization.

As PGD gains popularity,
many defense mechanisms
rely on it to evaluate their robustness.
Unfortunately,
AutoAttack (AA)~\cite{croce20aa}
finds that many of the defenses
may inadvertently break PGD-based attacks,
which result in drastic overestimation of their robustness,
and proposes to combine an ensemble of diverse attacks
to minimize robustness overestimation.
LAFEAT~\cite{lafeat}
learns to leverage intermediate layers of the DNN,
and shows that attacking multiple layers
can produce stronger attacks,
but unfortunately
it cannot be applied to ensemble defenses.
Adaptive Auto Attack (\aaa)~\cite{ye2022aaa}
improves attack success rates
by using the gradient directions
to prescribe a more effective initial random perturbation.
As defenders
may design mechanisms
to circumvent existing attacks,
Adaptive attacks~\cite{tramer2020adaptive}
manually tailor specific attack strategies
for an extensive set of defenses.
Finally,
Composite Adversarial Attacks (CAA)~\cite{mao2021caa}
further combine a large selection
of attack methods,
and use a genetic algorithm
to learn an optimal attacking sequence.
In comparison,
\attack{}
is a unified approach
which uses only one attack algorithm,
does not require
a compute-intensive learning procedure,
and yet it still achieves
fast and SOTA estimation
of ensemble robustness.

\subsection{Defending Against Adversarial Attacks}

Defending against adversarial attacks
can be defined as a saddle-point problem
to minimize the training loss
on adversarial examples~\cite{madry18}
with samples \( (\x, y) \) drawn from the training set:
\begin{equation}
    \min_{\allw} \expect[\parens{\x, y}]{
        \max_{\xadv \in \advset}
        \loss\parens{ f\parens\xadv, y }
    },
\end{equation}
where \( \loss \) is the training loss,
\eg{}, the SCE loss.
A direct optimization-based approach
to approximately
solving the above problem
is \emph{adversarial training}~\cite{madry18},
\ie{}, to train the DNN model
with its own adversarial examples.
Training DNNs to be robust
is, however, a challenging endeavor.
First,
it may be much more compute intensive
as training examples
are typically generated with PGD~\cite{madry18},
requiring a few forward/backward passes
of the DNN\@.
Second,
to avoid overfitting,
it requires stopping training early,
a much larger size of the training set~\cite{carmon19},
and using improved data augmentation~\cite{rebuffi2021data}
or generated data~\cite{gowal2021generated}.
Thirdly,
as noted
by other literatures~\cite{croce20aa,lafeat},
currently no other design choices
can rival the robustness provided by adversarial training,
and notably,
many defense strategies
are considered harmful
to model robustness~\cite{tramer2020adaptive}.
Finally,
the resulting models often
cannot achieve high clean accuracy~\cite{tsipras2018robustness}.

\subsection{Ensemble Defenses \& Ensemble-forming Strategies}

Ensemble-based defense techniques
may pave an alternative path
to address the challenges of adversarial robustness,
as they could potentially work around
the above limitations of adversarial training.
Adopting the theme
of minimizing adversarial example transferability
across sub-models,
each ensemble defense
proposed unique solutions.
ADP~\cite{pang2019adp}
increases the orthogonality
of non-maximal class logits among sub-models
to encourage diversity.
GAL~\cite{kariyappa2019gal}
minimizes a gradient alignment loss,
which directly reduces the cosine-similarity
between sub-models.
Building on top of this,
TRS~\cite{yang2021trs}
further regularizes the smoothness of the loss function,
as gradient orthogonality with smoothness
may further diversify sub-models.
Dverge~\cite{yang2020dverge}
instead uses the adversarial examples
of a sub-model to train another sub-model,
thus lowering transferability.
Ensemble defenses are also particularly interesting,
as they are the last line of defense
against even the strongest existing white-box attacks
without resorting to adversarial training,
showing a certain degree of robustness.


Besides the above mechanisms
for training a successful ensemble defense,
there exists different ways
to combine sub-model predictions.
Let us assume
that an ensemble defense
trains \( M \) sub-models,
\( \f{m}: \inputset \to \outputset \)
for \( m \in [1:M] \),
an ensemble
\( \f{\E}: \inputset \to \outputset \)
thus forms a final classification result
by combining individual decisions
from the sub-models,
namely:
\begin{equation}
    \textstyle \f{\E}(\x) =
    \frac1M \ssum_{m\in[1:M]} \ensop\parens{\f{m}\parens{\x}},
\end{equation}
where \( \ensop \) is the ensemble-forming operator.
In this paper,
we investigate
\( \ensop \in \braces{\softmax, \wta, \id} \),
where the potential choices
respectively denoting forming an ensemble
from sub-model outputs \( \f{m}\parens{\x} \)
by either summing predicted probabilities
(evaluated with \( \softmax \)),
or majority votes
(using \( \wta \), the winner-take-all operator),
or simply summing the logits
(with \( \id \), the identity operator).
Defending ensemble methods~\cite{
    pang2019adp,kariyappa2019gal,yang2020dverge,yang2021trs}
tested in this paper
all employed the \( \softmax \)-based strategy
to report their robustness.
Methods that are exceptions
to these options exist,
for instance,
ECOC~\cite{verma2019ecc}
allows sub-models
to produce binary predictions,
and use error correcting codes
based on the Hamming distance
to combine the predictions
into classification outputs.
This approach
is unfortunately not robust,
and the added complexity
is error-prone
and may harm robustness~\cite{tramer2020adaptive}.

Moreover,
as the voting (\( \wta \)) strategy
is non-differentiable,
an attacker can soften it approximately
using a softmax operation
with temperature \( \tau \),
where we used \( \tau = 0.1 \) universally:
\begin{equation}
    \textstyle
    \softwta_{\tau}\parens{\x}
    \triangleq \softmax\parens*{\x / \tau}.
    \label{eq:softwta}
\end{equation}

%% file: method.tex
\section{%
    The Model-Reweighing Attack (\attack)
}\label{sec:method}

\subsection{%
    Problem Formulation \& High-Level Overview
}\label{sec:method:overview}

As discussed in \Cref{sec:intro},
existing ensemble defenses
may obfuscate gradients
with the ensemble-forming mode
and gradient diversification,
such that the final loss of the ensemble model
can no longer provide effective signals
for gradient descent.
It is therefore
desirable to find an alternative \( \loss \)
to the original SCE loss \( \sceloss \)
on the ensemble,
such that for a given number of iterations \( I \),
the original \( \sceloss \) loss can be maximized:
\begin{equation}
    \textstyle
    \begin{aligned}
    &{\max}_{\loss} \sceloss\parens{\xadv_I, y}
    \quad \text{where} \quad
    \xadv_{0} = \project\parens{\x + \m},\, \\
    &\quad\xadv_{i + 1} = \mathrm{PGD}\parens{
        \loss\parens{
            \f{E}\parens{\xadv_i}, \f{[1:M]}\parens{\xadv_i},
        y}
    },
    \end{aligned}
\end{equation}
and \( \mathrm{PGD}\parens{\cdot} \)
denotes a PGD step
along the gradient of loss function \(
    \loss\parens{
        \f\E\parens{\xadv_i},
        \f{[1:M]}\parens{\xadv_i},
    y}
\),
which not only takes the ensemble predictions
\( f_E\parens{\xadv_i} \) as input,
but can further utilize sub-model predictions
\( \f{[1:M]} \)
to guide the PGD iterations.
The challenge at hand
is, therefore,
to find a suitable \( \loss \)
which can generate attacks
on ensemble defenses efficiently and effectively.


\input{figures/overview}

\attack{} aims to provide
a potential optimization route
towards the above problem formulation.
Namely,
in addition to the original output
of the ensemble \( \f\E\parens{\xadv} \),
we leverage the sub-model predictions
\( \f{[1:M]}\parens{\xadv} \)
to facilitate the optimization.
By way of illustration,
\Cref{fig:method:overview}
shows a high-level overview
of the model-reweighing attack,
where we compliment the ensemble loss,
with a newly added sub-model reweighing loss
\( \attackloss \),
as an auxiliary attack vector
alongside the original objective.
Not only can the new loss
bypass the ensemble-forming strategy
to work around its obfuscated gradients,
but it further exploits information
present in the individual sub-model
and ensemble predictions
to steer the direction of adversarial example synthesis.

\subsection{%
    Adaptive Sub-model Importance
}\label{sec:method:weight}

Before we begin,
assume that
\( \logit{m} \triangleq f_m \parens{\x} \)
represents the \( \ordinal{m} \)
sub-model output,
and let \( \logit{m}_t \)
denote the corresponding logit of label \( t \).
We define the difference of logits (DL)~\cite{carlini17}
\( \dl{m} \triangleq \logit{m}_y - \logit{m}_\hy \),
which is the difference
between the predictions
of the ground truth \( \logit{m}_y \)
and the maximum of the remaining classes
\(
    \logit{m}_\hy
    \triangleq
    \max_{i \in \classset / y} \logit{m}_i
\),
where \( \classset / y \)
is the set of all class labels except \( y \).
Similarly,
we let \( \logit{\E} \triangleq f_\E\parens{\x} \),
and \( \logit{\E}_t \) and \( \dl{\E} \)
be the respective variants of the ensemble prediction.
It is notable that
a successful attack happens
when \( \dl\E < 0 \),
and similarly \( \dl{m} < 0 \)
means the \( \ordinal{m} \) sub-model
is producing incorrect classification.

Ensemble defenses
tend to diversify sub-model gradients,
for instance, ADP~\cite{pang2019adp}
minimizes the cosine-similarity \(
    \angles{
        \nabla \ell\parens{\logit{a}},
        \nabla \ell\parens{\logit{b}}
    }
\) among each loss function gradient pairs
of sub-models \(
    \ell\parens{\logit{a}}
\) and \(
    \ell\parens{\logit{b}}
\).
Their intuition
is that it may lower transferability
among these sub-models,
such that attacks
with the overall gradient of the ensemble,
\ie{}, \(
    \nabla \ell\parens{\logit\E}
    = \frac1M \sum_{m \in [1:M]}
        \nabla \ell\parens{\logit{m}}
\),
are becoming less effective
in misleading all sub-models simultaneously,
as individual gradients in \(
    \nabla \ell\parens{\logit{m}}
\) are encouraged to be orthogonal to each other.
To this end,
we propose to reweigh the importance of sub-models,
by instead considering the modified gradient:
\begin{equation}
    \textstyle
    \widehat{\nabla \ell\parens{\logit\E}}
    = \frac1M \sum_{m \in [1:M]}
        { \imp{m}\parens{\logit{m}} }
        \nabla \ell\parens{\logit{m}},
\end{equation}
where
\( \imp{m} \)
assigns weights to important sub-models
to contribute more heavily
to the attack gradient.

While the adversarial examples of the ensemble
could present a challenge to discover,
individual sub-models
are weak defenders
which can be easily defeated.
Based on this property,
we propose to weigh sub-models importance
with the rate of change in \( \dl{\E} \)
\wrt{} that of \( \dl{m} \),
\ie{}, sub-models would be given higher weights
if attacking it would bring a significant change
to the ensemble's prediction.
Following this idea,
for all ensemble-forming strategies (softmax, voting, logits),
we rewrite \( \dl{\E} \)
as a function of \( \dl{m} \),
where the term below
can become a function of \( \dl{m} \):
\begin{equation}
    \textstyle
    \begin{aligned}
    \dl{\E} = \ensop\parens{\logit{m}}_\y -
    \ensop\parens{\logit{m}}_\hy
    &= \ensop\parens{\logit{m} - \logit{m}_\hy}_\y -
      \ensop\parens{\logit{m} - \logit{m}_\hy}_\hy \\
    &= \ensop\parens{\dl{m}, \cdots}_\y -
      \ensop\parens{\dl{m}, \cdots}_\hy
    \triangleq h_m\parens{\dl{m}}.
    \end{aligned}
\end{equation}
The weights are thus defined as follows:
\begin{equation}
    \textstyle
    \imp{m}\parens{\logit{m}}
    = \frac{\partial \dl{\E}\parens{\dl{m}}}{\partial \dl{m}}
    = \frac\partial{\partial\dl{m}}\parens*{
        {\frac1M} {\ssum}_{m\in[1:M]}
            \frac{\partial h_m\parens{\dl{m}}}{\partial \dl{m}}
    }
    = \frac1M \frac{\partial h_m\parens{\dl{m}}}{\partial \dl{m}}.
\end{equation}
While it is possible
to compute the weights using gradient back-propagation,
we can simply derive the following closed-form solution
of the weights
for each of the three ensemble-forming strategies.
For \( \wta \),
we use the softened version of \( \wta \)
as defined in~\eqref{eq:softwta}
and can derive the weights as follows:
\begin{equation}
    \newcommand{\sv}{\mathbf{s}}
    \imp{m}\parens{\logit{m}} =
    \indicator\bracks{\dl{m} > 0} \cdot
    \detach\parens*{
        {\textstyle\frac1{\tau M}}
        \sv_\hy \parens*{1 + \sv_\y - \sv_\hy}
    },
    \quad \text{where}\,\,
    \sv = \softmax\parens*{\nicefrac{\logit{m}}\tau}.
    \label{eq:weight}
\end{equation}
Here \( \indicator\bracks{\dl{m} > 0} \)
is the indicator function
that equals 1 if \( \dl{m} > 0 \),
or 0 otherwise,
effectively stopping the attack
on the \( \ordinal{m} \) sub-model upon success,
and the \( \detach \) operator
admits no backward propagation to its input.
In the case of using sums
of sub-model softmax outputs
to form an ensemble decision,
\ie{}, \( \ensop = \softmax \),
it is a special case of \( \softwta_\tau \)
where the temperature coefficient
can be fixed at \( \tau = 1 \).
Finally,
when \( \ensop = \id \),
\ie{}, forming ensembles by summing logits,
\( \imp{m}\parens{\logit{m}} \)
simply reduces to \( \indicator\bracks{\dl{m} > 0} \)
for the \( \ordinal{m} \) sub-model.
\input{algorithm}

\subsection{The \attack{} Loss}\label{sec:method:loss}

For reference,
defenses mechanisms we examine
in this paper
aim to find \( \xadv \)
which maximizes the SCE loss
\(
    \sceloss\parens*{
        \logit{\E}, \y
    },
\)
to evaluate the ensemble robustness.
The \attack{} loss improves this further
by proposing two additional modifications
to the untargeted loss function used to attack ensembles:
\begin{equation}
    \attackloss
        \parens{\logit{1:M}, \logit{\E}, y}
    \triangleq
    \sceloss\parens*{
        \beta \scenorm\parens*{
            \ssum_{m \in [1:M]}
            \imp{m}\parens{\logit{m}} \cdot \logit{m}
        }
        + (1 - \beta) \scenorm\parens*{\logit\E}, y
    }.
    \label{eq:attackloss}
\end{equation}
First,
it additionally introduces a sum
of the \( \imp{m} \)-weighted variant
of sub-model logits,
in order to expose sub-model logits
with adaptive reweighing described in \Cref{sec:method:weight}.
Second,
\( \beta \)
interpolates the importance
of the newly added auxiliary logits
and the original ensemble logits.  
Finally,
inspired by the effective surrogate loss
in~\cite{lafeat},
it further normalizes the logits
by their respective DL using:
\begin{equation}
    \scenorm\parens{\z} \triangleq
        \indicator\bracks*{\z_\y - \z_\hy > 0}
        \cdot
        {\z} / {\detach\parens{\z_\y - \z_\hy}}.
    \label{eq:scenorm}
\end{equation}

Finally, the targeted variant
of the \attack{} loss simply replaces \( y \) with \( t \)
where \( t \) is the intended target.

\subsection{Improving the State-of-the-art}

While the new \( \attackloss \) loss
is highly effective against
ensemble defenses we test in this paper,
we strive for further advances
in \attack's ability
to generate faster and better adversarial examples.
Inspired by recent publications,
we borrow ideas
from related adversarial attack tactics,
which includes
adopting a cosine step-size schedule~\cite{ye2022aaa},
momentum~\cite{dong18momentum,croce20aa},
random restarts~\cite{tramer2020adaptive}
and multiple target attacks~\cite{croce20aa,tramer2020adaptive}.
We provide the overall algorithm
in \Cref{alg:overview},
which computes
an adversarial image \( \xadv_I \)
as return,
by taking as input
the sub-models \( \f{[1:M]} \),
natural image \( \x \),
ground truth label \( \y \),
\( \beta \) to interpolate between
the auxiliary logits and the original,
\( \tau \) controls the temperature,
momentum \( \mu = 0.75 \)
following~\cite{lafeat,croce20aa},
\( \epsilon \) perturbation bound,
and finally the maximum number of iterations \( I \).

%% file: figures/overview.tex
\begin{figure*}[!ht]
    \centering%
    \includegraphics[scale=0.5]{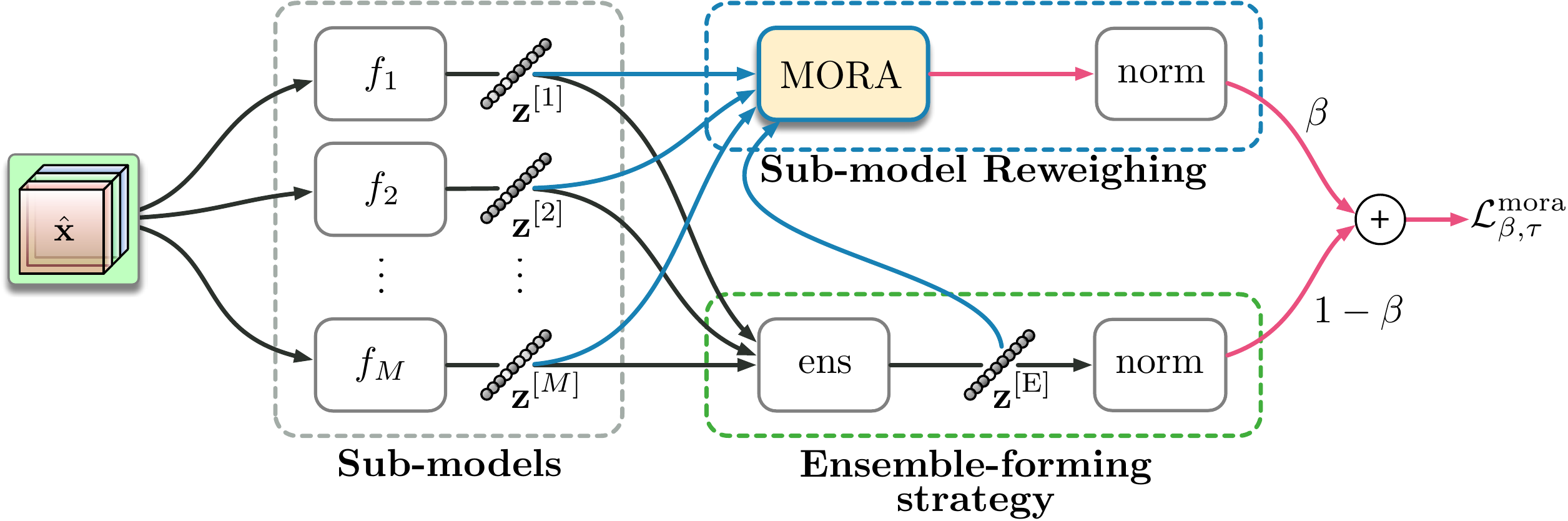}%
    \caption{%
        A high-level overview of \attack.
        Existing attack methods
        focus on maximizing the \( \sceloss \)
        of the ensemble predictions \( \logit\E \).
        \attack{} further
        introduces a model-reweighing mechanism,
        which takes as input
        sub-models predictions
        \( {\logit{1:M}} \),
        and the ensemble output \( \logit\E \),
        forming a combined loss \( \attackloss \).
        \attack{} generates adversarial examples
        by maximizing the resulting loss.
    }\label{fig:method:overview}
\end{figure*}

%% file: algorithm.tex
\begin{algorithm}[t]
\caption{%
    The \attack{} white-box robust evaluation for ensemble defenses.
}\label{alg:overview}
\algnewcommand{\IfThen}[2]{
    \State{\algorithmicif\ {#1}\ \algorithmicthen\ {#2}}}
\algnewcommand{\IfThenElse}[3]{
    \State{\algorithmicif\ {#1}\ %
    \algorithmicthen\ {#2}\ \algorithmicelse\ {#3}}}
\newcommand{\algcmt}{\algorithmiccomment}
\newcommand{\submodelset}{\f{\bracks{1:M}}}
\begin{algorithmic}[1]
    \Function{\tt\attack\_\,Attack}{$
        \submodelset,
        \x, \y, \beta, \tau, \momentum, \epsilon, I
    $}
        \State{\(
            \xadv_0 \gets \project\left(
                \x + \mathbf{u}
            \right),\,\text{where}\,
            \mathbf{u} \sim \uniform{-\epsilon, \epsilon}
         \)}
        \algcmt{Random init}
        \State{\( \m_0 \gets 0 \)}
        \For{\( i \in \bracks{0:I-1} \)}
            \State{\(
                \logit{m} \gets \f{m}\parens{\xadv_i}
             \) for all \(  {m} \in [1:M]  \)}
            \algcmt{Sub-model predictions}
            \State{\(
                \logit{\E} \gets
                    \frac1M \ssum_{m \in [1:M]}
                    \ensop\parens{\logit{m}}
             \)}
            \algcmt{Ensemble prediction}
            \State{\( \dl\E \gets \logit\E_\y - \logit\E_\hy \)}
            \IfThen{\( \dl\E \leq 0 \)}{\Return{\( \xadv_i \)}}
            \algcmt{Successful attack}
            \State{\(
                \bm{g}_{i + 1} \gets
                    \sign\parens{\nabla_{\xadv_i}
                        \attackloss\parens{
                        \logit{1:M}, \logit\E, y
                    }
                }
             \)}
            \algcmt{Sign-gradient with the \attack{} loss}
            \State{\(
                \stepsize \gets \epsilon \parens*{
                    1+\cos \parens*{ \nicefrac{i\pi}{I} }
                }
             \)}
            \algcmt{Cosine step-size schedule}
            \State{\(
                \m_{i + 1} \gets \project\left(
                    \m_i + \stepsize \bm{g}_{i + 1}
                \right)
             \)}
            \algcmt{Iterative update}
            \State{\(
                \xadv_{i + 1} \gets \project\left(
                    \xadv_i + {}
                    \momentum \left( \m_{i+1} - \xadv_i \right) +
                    (1 - \momentum) \left( \xadv_i - \xadv_{i - 1} \right)
                \right)
             \)}
            \algcmt{\ldots with momentum}
        \EndFor{}
        \State{\Return{\( \xadv_I \)}}
        \algcmt{Give up after \( I \) iterations}
    \EndFunction%
\end{algorithmic}
\end{algorithm}%

%% file: results.tex
\section{Experimental Results}\label{sec:results}

We compare \attack{} against SOTA attacks
for a wide range of ensemble defenses
under three ensemble-forming strategies
(softmax, voting, and logits).
We use pre-trained ResNet-20~\cite{resnet} models
from ADP~\cite{pang2019adp},
Dverge~\cite{yang2020dverge},
GAL~\cite{kariyappa2019gal},
and reproduced TRS~\cite{yang2021trs}
using the same architecture
with official source code,
as pre-trained models were unavailable.
Our robustness evaluation
considers the \( \linf \) white-box attacks
on the \cifarx{} test set~\cite{cifar},
with perturbation \( \epsilon = 0.01 \)
unless specified.
The full comparison results
can be found in \Cref{tab:compare:cifarx};
larger \( \epsilon \) comparisons,
and similar results on \cifarc{} models
are in \Cref{app:results}.
We provide our key observations below.
\input{tables/cifar10}

\textbf{%
    Traditional attacks 
    may fail to break through gradient obfuscation.}
We reproduce two traditional white-box attacks,
\ie{}, projected gradient descent (PGD)~\cite{madry18}
and C\&W~\cite{carlini17}
with 5 random restarts,
each with a maximum of 100 iterations,
giving a total of 500 iterations.
PGD uses a fixed step size of \( \epsilon/4 \).
For a fair comparison,
\attack{} with 500 iterations
sweeps \( \beta \in \braces{0, 0.25, 0.5, 0.75, 1} \),
with each \( \beta \) up to 100 iterations.
Even with a 500 iteration budget,
it is clear that PGD and C\&W
may substantially overestimate robustness,
especially when tested
under the softmax and voting ensemble-forming options,
and \attack{} can work around this obstacle
thanks to its attacks on sub-models.

\textbf{%
    Diversified gradients
    can hamper even integrated attacks with large arsenals.}
Moreover, we test the defenses
against recent integrated attacks
with SOTA baselines on robustness evaluation,
namely Adaptive Auto Attack (\aaa)~\cite{ye2022aaa},
AutoAttack (AA)~\cite{croce20aa},
and Composite Adversarial Attacks (CAA)~\cite{mao2021caa},
which comprise large arsenals of various attack strategies.
We reproduce CAA following~\cite{mao2021caa}
to search for the attack policy
before evaluating the defending models.
Note that
its computational complexity
is thus much higher than the other attacks,
but we only report its test-time complexity.
In particular,
while they enjoyed
much higher success rates than PGD and \cw,
some defenses render their tactics ineffective.
We observe, \eg, sizeable robustness overestimation
on ADP~\cite{pang2019adp} under softmax and voting,
which explicitly diversifies sub-model gradients.
As \attack{} can dynamically
re-adjust sub-model importance
\wrt{} their ``ease-of-attack'',
it performs substantially better
with much fewer iterations.
In addition to the earlier 500 iterations,
the multi-targeted \attackmt{}
targets the remaining 9 class labels
with 100 iterations for each label
and \( \beta \) fixed at \( 0.5 \).
Others also use multi-targeted attacks
along with respective tactics.

\textbf{%
    Robustness of most sub-models
    \vs{} robustness of ensemble.}
We find that robustness
of a majority of sub-models
(fooling \( \nicefrac38 \) for softmax
 and \( \nicefrac28 \) for logits)
usually do not translate
to the overall robustness
of the ensemble
(\Cref{fig:num_models/softmax,fig:num_models/logits}).
As voting requires breaking
\( \nicefrac12 \) sub-models simultaneously
(\Cref{fig:num_models/argmax}),
it is perceivable that using voting
may give rise to a higher overall robustness.
Yet surprisingly,
for most defending ensembles,
voting performs worse than softmax and logits.

\textbf{%
    Ensemble-forming strategies
    may give a false sense of security.}
On one hand,
softmax and voting strategies
exhibit substantially larger overestimated robustness
(up to 40\%)
than logits.
On the other hand,
in stark contrast
to the proposed use of softmax
from~\cite{
    kariyappa2019gal,pang2019adp,yang2020dverge,yang2021trs},
we find summing by logits
can form ensembles
that are notably more robust
than the other two
(\Cref{fig:epsilon/dverge}),
while attackers only needs
to successfully deceive a few sub-models
(referring back to~\Cref{fig:num_models/logits}).
\input{figures/num_models}
\input{figures/convergence}

\textbf{%
    Up to \( 60\times \) faster convergence
} under 500 iterations.
\Cref{fig:results:convergence}
compares the convergence speed
of \attack{} against AA losses, \cw{}, and PGD
on defending ensembles.
\attack{}
converges substantially faster
than the other attacks,
using only up to 31 steps
to match AA losses with \( 500 \) iterations.

\textbf{%
    Ensemble defense mechanisms
    may be at odds with robustness.}
In \Cref{tab:compare:cifarx:at},
we compare respective attacks
on adversarially trained Dverge models.
To our surprise,
forming larger ensembles
is actually harmful
to the robustness of ensemble.
\input{tables/adv_ensemble}


\textbf{%
    Additional results, ablation, and sensitivity analyses.
}
Due to the page limit,
we provide full results
of relevant figures
in \Cref{app:results},
note that the above key observations
still hold true for all ensemble defenses we test
under different ensemble-forming strategies
and \( \epsilon \) perturbation bounds.
In addition,
we provide extensive ablation study
on the design choices we made,
and sensitivity analysis
on the temperature constant \( \tau \).

%% file: tables/cifar10.tex
\begin{table*}[t]
\centering%
\caption{%
    Comparing accuracies among
    iterative methods~\cite{madry18,carlini17},
    learned attacks (\textbf{CAA}~\cite{mao2021caa}),
    AutoAttack (\textbf{AA})~\cite{croce20aa},
    adaptive auto attack
    (\( \textbf{A}^\mathrm{3} \))~\cite{ye2022aaa},
    and \attack{}
    across various ensemble defense strategies
    under 3 ensembling modes
    (softmax, voting and logits),
    and \( \epsilon = 0.01 \).
    The ``Complexity'' row
    shows the worst-case complexity
    in iteration counts.
    The ``\( \bm\Delta \)'' column
    shows the accuracy overestimation
    from self-reported/reproduced
    ``\textbf{Nominal}'' values
    to \attackmt{}.
    Baselines with \textdagger{}
    are reproduced with source code.
    All results are re-run 5 times
    and within \( \pm 0.05\% \) standard deviation.
    \vspace{-5pt}
}\label{tab:compare:cifarx}
\adjustbox{width=\textwidth}{%
\begin{tabular}{cc|cc|ccc|cccc|c}
\toprule
\tbox{\textbf{Defense} \\ Complexity}
    & \textbf{\#}
    & \tbox{\textbf{Clean} \\ 1}
    & \tbox{\textbf{Nominal} \\ \tna}
    & \tbox{\textbf{PGD} \\ 500} & \tbox{\textbf{CW} \\500}
    & \tbox{\textbf{\attack} \\500}
    & \tbox{\( \textbf{A}^\mathrm{3} \) \\12k}
    & \tbox{\textbf{AA} \\4.9k} & \tbox{\textbf{CAA} \\1.8k}
    & \tbox{\textbf{\attackmt} \\1.4k}
    & \( \bm\Delta \) \\
\midrule
\multicolumn{12}{c}{\textbf{Softmax}} \\
\midrule
\multirow{3}{*}{ADP}
    & 3 & 92.88 & 29.12
        &  5.98 &  7.72 &  0.59
        &  2.12 &  0.98 &  3.34
        & \textbf{ 0.34} &  28.78 \\
    & 5 & 93.34 & 25.14
        &  7.10 &  8.70 &  0.97
        &  3.62 &  2.18 &  4.25
        & \textbf{ 0.67} &  24.47 \\
    & 8 & 93.48 & 20.20
        &  9.22 &  9.59 &  1.70
        &  4.84 &  3.94 &  6.04
        & \textbf{ 1.32} &  18.88 \\
\midrule
\multirow{3}{*}{Dverge}
    & 3 & 91.99 & 47.42
        & 44.49 & 40.17 & 25.77
        & 33.36 & 30.58 & 32.98
        & \textbf{25.26} & 22.16 \\
    & 5 & 92.38 & 55.72
        & 54.61 & 52.83 & 40.02
        & 48.41 & 43.29 & 46.65
        & \textbf{39.50} & 16.22 \\
    & 8 & 91.65 & 59.63
        & 59.13 & 58.25 & 55.68
        & 57.29 & 56.71 & 56.89
        & \textbf{55.57} &  4.06 \\
\midrule
\multirow{3}{*}{GAL}
    & 3 & 89.41 & 19.48
        &  8.13 & 11.57 &  0.67
        &  0.70 &  0.85 &  1.00
        & \textbf{ 0.51} & 18.97 \\
    & 5 & 90.93 & 41.38
        & 37.59 & 35.52 & 17.45
        & 26.94 & 23.90 & 25.11
        & \textbf{16.05} & 25.33 \\
    & 8 & 92.45 & 56.31
        & 53.39 & 52.56 & 28.71
        & 36.51 & 37.46 & 35.30
        & \textbf{27.44} & 28.87 \\
\midrule
\multirow{3}{*}{TRS\supdagger}
    & 3 & 70.02 & 19.71
        & 14.01 & 10.87 &  8.11
        &  8.72 &  8.46 &  9.75
        & \textbf{ 7.60} &  12.11 \\
    & 5 & 69.00 & 23.17
        & 15.91 & 15.28 & 12.67
        & 13.22 & 13.20 & 13.78
        & \textbf{12.47} &  10.70 \\
    & 8 & 73.01 & 23.64
        & 18.02 & 17.59 & 15.90
        & 16.22 & 16.51 & 16.73
        & \textbf{15.64} &  8.00 \\
\midrule
\multicolumn{12}{c}{\textbf{Voting}} \\
\midrule
\multirow{3}{*}{ADP}
    & 3 & 91.84 & 41.62\supdagger
        &  9.32 & 11.84 &  0.64
        &  3.06 &  6.13 &  8.29
        & \textbf{ 0.29} &  41.33 \\
    & 5 & 93.13 & 40.29\supdagger
        & 12.42 & 12.05 &  1.17
        &  6.03 & 10.13 &  0.67
        & \textbf{ 0.62} & 39.67 \\
    & 8 & 93.28 & 30.10\supdagger
        & 12.53 & 10.50 &  3.16
        &  6.11 &  9.21 &  1.69
        & \textbf{ 1.65} &  28.45 \\
\midrule
\multirow{3}{*}{Dverge}
    & 3 & 91.72 & 39.05\supdagger
        & 31.48 & 28.00 & 23.57
        & 24.95 & 24.98 & 27.65
        & \textbf{22.91} &  16.14 \\
    & 5 & 92.18 & 49.36\supdagger
        & 44.28 & 42.28 & 35.06
        & 39.15 & 39.20 & 40.85
        & \textbf{34.46} &  14.90 \\
    & 8 & 91.58 & 56.85\supdagger
        & 53.72 & 52.35 & 47.12
        & 50.58 & 50.04 & 51.15
        & \textbf{46.10} &  10.75 \\
\midrule
\multirow{3}{*}{GAL}
    & 3 & 89.09 & 21.48\supdagger
        &  5.85 &  7.64 &  0.87
        &  0.71 &  0.56 &  0.78
        & \textbf{ 0.35} &  21.13 \\
    & 5 & 90.77 & 37.32\supdagger
        & 29.33 & 27.62 & 12.96
        & 18.55 & 20.82 & 22.17
        & \textbf{12.25} & 25.07 \\
    & 8 & 92.37 & 55.39\supdagger
        & 49.56 & 48.02 & 21.66
        & 30.35 & 31.39 & 30.93
        & \textbf{20.16} & 35.23 \\
\midrule
\multirow{3}{*}{TRS\supdagger}
    & 3 & 68.95 & 13.79
        & 10.19 &  8.71 &  5.73
        & 11.89 &  6.69 &  8.08
        & \textbf{ 5.44} &  8.35 \\
    & 5 & 68.31 & 15.36
        & 12.71 & 11.88 &  8.82
        & 10.08 & 10.30 & 11.21
        & \textbf{ 8.38} &  6.98 \\
    & 8 & 72.05 & 17.00
        & 14.57 & 13.48 & 11.39
        & 11.99 & 11.85 & 12.80
        & \textbf{10.69} & 6.31 \\
\midrule
\multicolumn{12}{c}{\textbf{Logits}} \\
\midrule
\multirow{3}{*}{ADP}
    & 3 & 92.86 & 3.44\supdagger
        &  0.87 &  2.05 &  0.48
        &  0.25 &  0.22 &  0.31
        & \textbf{ 0.21} & 3.23 \\
    & 5 & 93.48 & 4.57\supdagger
        &  1.97 &  4.24 &  1.12
        &  1.00 &  0.97 &  1.09
        & \textbf{ 0.89} & 3.68 \\
    & 8 & 93.38 & 5.39\supdagger
        &  3.57 &  4.77 &  2.13
        &  2.20 &  2.05 &  2.11
        & \textbf{ 1.93} &  3.46 \\
\midrule
\multirow{3}{*}{Dverge}
    & 3 & 92.19 & 38.31\supdagger
        & 37.99 & 38.60 & 36.89
        & 36.94 & 36.96 & 37.07
        & \textbf{36.84} &  1.47 \\
    & 5 & 92.28 & 50.77\supdagger
        & 50.57 & 51.28 & 49.65
        & 49.72 & 49.66 & 49.75
        & \textbf{49.59} &  1.18 \\
    & 8 & 91.73 & 61.06\supdagger
        & 60.95 & 61.51 & 60.52
        & 60.59 & 60.52 & 60.55
        & \textbf{60.49} &  0.57 \\
\midrule
\multirow{3}{*}{GAL}
    & 3 & 89.50 & 15.47\supdagger
        & 10.01 & 10.53 &  0.52
        &  \textbf{0.02} &  \textbf{0.02} &  0.08
        &  0.03 & 15.44 \\
    & 5 & 90.93 & 36.36\supdagger
        & 33.97 & 35.14 & 22.24
        & 33.43 & 20.24 & 21.66
        & \textbf{19.40} & 16.96 \\
    & 8 & 92.54 & 56.08\supdagger
        & 53.67 & 54.69 & 31.52
        & 40.90 & 30.89 & 31.17
        & \textbf{30.66} & 25.42 \\
\midrule
\multirow{3}{*}{TRS\supdagger}
    & 3 & 69.72 & 13.31
        & 13.06 & 13.80 & 12.11
        & 12.13 & 12.16 & 12.21
        & \textbf{12.07} &  1.24 \\
    & 5 & 68.90 & 16.89
        & 16.65 & 17.34 & 15.88
        & 15.86 & 15.90 & 15.95
        & \textbf{15.82} &  1.07 \\
    & 8 & 72.24 & 19.40
        & 19.20 & 19.67 & 18.20
        & 18.18 & 18.27 & 18.34
        & \textbf{18.17} &  1.23 \\
\bottomrule
\vspace{-20pt}
\end{tabular}
}\end{table*}

%% file: figures/num_models.tex
\begin{figure}[!b]
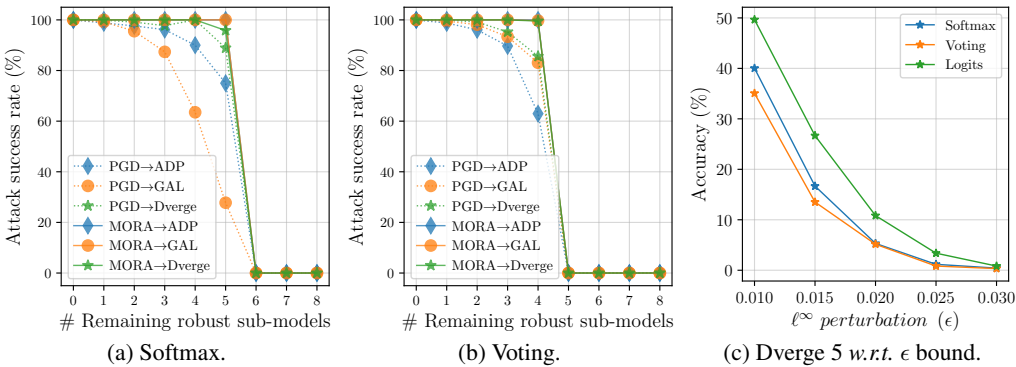

    \centering%
    \subplot{num_models/softmax}{Softmax}
    \subplot{num_models/argmax}{Voting}
    \subplot{epsilon/dverge}{%
        Dverge 5 \wrt{} \( \epsilon \) bound}
    \caption{%
        (\subref{fig:num_models/softmax},
         \subref{fig:num_models/argmax})
        \attack{} can successfully fool
        ensemble-forming methods
        (ADP~\cite{pang2019adp}, GAL~\cite{kariyappa2019gal}
         and Dverge~\cite{yang2020dverge})
        even with the majority of their sub-models
        giving correct outputs
        under softmax and logits (\Cref{fig:num_models/logits}).
        While voting
        requires \( \geq \nicefrac12 \) sub-models
        to be incorrect,
        it is unfortunately the least robust option
        in all defenses.
        ``\( A \rightarrow B \)''
        means using \( A \) to attack \( B \)
        for up to 100 iterations.
        (\subref{fig:epsilon/dverge})
        Dverge with 5 sub-models \wrt{}
        the \( \epsilon \) bound on \( \linf \) perturbation.
        Contrary to existing literatures,
        we find logits to be the most robust option
        of the three ensemble-forming strategies.
    }\label{fig:results:num_models}\vspace{-5pt}
\end{figure}

%% file: figures/convergence.tex
\begin{figure}[!ht]
    \centering%
    \subplot{convergence/dverge_3_softmax}{Dverge 3}
    \subplot{convergence/dverge_8_softmax}{Dverge 8}
    \subplot{convergence/gal_8_softmax}{GAL 8}
    \caption{%
        (\subref{fig:convergence/dverge_3_softmax},
         \subref{fig:convergence/dverge_8_softmax},
         \subref{fig:convergence/gal_8_softmax})
        Comparing the convergence speed
        of \attack{}
        against a variant of \attack{}
        without sub-model reweighing,
        \cw{}, PGD and AA losses (APGD-\{DLR,CE\})
        on Dverge and GAL
        with softmax.
        While ``No reweigh''
        converge faster initially,
        it struggles to improve after 10 iterations;
        in contrast, adaptive reweighing (\attack)
        continues to converge at a much faster rate.
        The horizontal and vertical axes
        respectively show the iteration count used,
        and the percentage of unsuccessful images remaining.
        We annotated the number of iterations
        for \attack{} to overtake all
        with 500 iterations.
    }\label{fig:results:convergence}
\end{figure}

%% file: tables/adv_ensemble.tex
\begin{table*}[ht]
\centering%
\caption{%
    Attacking adversarially trained
    Dverge~\cite{yang2020dverge} models
    under the same setting
    as \Cref{tab:compare:cifarx},
    except we let \( \epsilon = 0.03 \).
    Notably, forming larger ensembles
    can actually be detrimental to robustness.
}\label{tab:compare:cifarx:at}
\adjustbox{width=\textwidth}{%
\begin{tabular}{cc|cc|ccc|cccc|c}
\toprule
\tbox{\textbf{Dverge} \\ Complexity}
    & \textbf{\#}
    & \tbox{\textbf{Clean} \\ 1}
    & \tbox{\textbf{Nominal} \\ \tna}
    & \tbox{\textbf{PGD} \\ 500} & \tbox{\textbf{CW} \\ 500}
    & \tbox{\textbf{\attack} \\ 500}
    & \tbox{\(\mathbf{A}^3\) \\12k}
    & \tbox{\textbf{AA} \\4.9k} & \tbox{\textbf{CAA} \\1.8k}
    & \tbox{\textbf{\attackmt} \\1.4k}
    & \( \bm\Delta \) \\
\midrule
\multirow{3}{*}{Softmax}
    & 3 & 83.78 & 45.09
        & 44.85 & 44.21 & 42.91
        & 42.66 & 42.70 & 42.69
        & \textbf{42.65} & 2.44 \\
    & 5 & 86.09 & 42.57
        & 42.40 & 42.51 & 41.05
        & \textbf{40.74} & 40.85 & 40.84
        & 40.75 & 1.79 \\
    & 8 & 86.69 & 40.80
        & 40.58 & 40.94 & 39.49
        & \textbf{39.33} & 39.40 & 39.39
        & 39.35 &  1.41 \\
\midrule
\multirow{3}{*}{Voting}
    & 3 & 83.67 & 59.13\supdagger
        & 55.35 & 55.85 & 38.59
        & 38.78 & 39.97 & 40.31
        & \textbf{38.24} &  20.89 \\
    & 5 & 86.05 & 47.72\supdagger
        & 44.32 & 45.18 & 36.19
        & 36.42 & 37.29 & 37.89
        & \textbf{36.01} &  11.71 \\
    & 8 & 86.54 & 38.93\supdagger
        & 37.16 & 38.17 & 34.20
        & 34.89 & 35.54 & 36.19
        & \textbf{34.03} &  4.90 \\
\midrule
\multirow{3}{*}{Logits}
    & 3 & 83.74 & 44.83\supdagger
        & 44.69 & 44.24 & 42.83
        & 42.66 & 42.70 & 42.70
        & \textbf{42.63} &  2.20 \\
    & 5 & 86.03 & 42.47\supdagger
        & 42.22 & 42.63 & 41.00
        & 40.91 & 40.92 & 40.92
        & \textbf{40.87} &  1.60 \\
    & 8 & 86.65 & 40.53\supdagger
        & 40.33 & 41.14 & 39.53
        & \textbf{39.43} & 39.50 & 39.47
        & \textbf{39.43} &  1.09 \\
\bottomrule
\end{tabular}
}\end{table*}

%% file: conclusion.tex
\section{Conclusions}

This paper identifies
severe robustness overestimation
in many ensemble defense techniques,
and further investigates
problem the robustness evaluation
under three ensemble-forming strategies.
To efficiently and accurately
evaluate the robustness of ensembles,
we introduce \attack{},
a new attack technique
which reweighs sub-model importance
adaptively by their respective ``ease-of-attack''
during attack iterations.
\attack{}
enjoys a much improved success rate
and convergence rate
compared with other SOTA attacks.
Moreover,
we found several surprising observations
related to ensemble defenses,
for instance,
(1) misleading a minority of sub-models
is sufficient to fool the ensemble,
(2) summing by logits
is the simplest yet most robust way
to form ensembles,
(3) with adversarial training,
ensemble defenses may actually
harm robustness, \etc{}
We hope the above observations
may help to guide future avenue
on ensemble defenses,
and provide a strong attack baseline
for potential approaches.
Finally, \attack{} is open source
with reproducible results and pre-trained models;
and we continually update
a leaderboard of ensemble defenses
under various attack strategies.

\section*{Acknowledgements}

This work is supported in part
by National Key R\&D Program of China
(\numero{2019YFB2102100}),
Key-Area Research and Development Program
of Guangdong Province (\numero{2020B010164003}),
Science and Technology Development Fund
of Macao S.A.R (FDCT)
under \numero{0015/2019/AKP},
and Shenzhen Science and Technology Innovation Commission
(\numero{JCYJ20190812160003719}).
This work was carried out
in part at SICC
which is supported by SKL-IOTSC,
University of Macau.

%% file: appendix.tex
\section{Additional Results}\label{app:results}

\subsection{%
    Larger Perturbations
}\label{app:results:epsilons}

\Cref{tab:compare:cifarx:0.02,tab:compare:cifarx:0.03}
compare the effectiveness
of iterative methods~\cite{madry18,carlini17},
learned attacks (\textbf{CAA}~\cite{mao2021caa}),
AutoAttack (\textbf{AA})~\cite{croce20aa},
\attack{} and \attackmt{}
across various ensemble defense strategies
with ResNet-20~\cite{resnet} sub-models
under 3 ensemble-forming modes
(softmax, voting and logits).
Here, \Cref{tab:compare:cifarx:0.02}
uses an \( \linf \) perturbation bound
\( \epsilon = 0.02 \),
whereas \Cref{tab:compare:cifarx:0.03}
uses \( \epsilon = 0.03 \).
All results are re-run 5 times
and within \( \pm 0.05\% \) standard deviation.
Note that under \( \epsilon = 0.02 \)
most defense methods
only retain up to \( 4\% \) robust accuracies.
Dverge~\cite{yang2020dverge} with 8 sub-models
however surprisingly has \( 23.31\% \)
robustness against the strongest \attackmt{} attack.
We believe this can be attributed
to the fact that
similarly to adversarial training,
Dverge diversifies sub-models
with adversarial examples from each other,
rather than via explicit regularization
as carried out by the other defenses.
It thus requires substantially
higher training cost than the others
(\( \sim 5\times \) ADP~\cite{pang2019adp}).%

\begin{figure}[ht]
    \centering
    \newcommand{\epsplot}[2]{
        \subplot{epsilon/#1_3}{#2 (3 sub-models)}
        \subplot{epsilon/#1_5}{#2 (5 sub-models)}
        \subplot{epsilon/#1_8}{#2 (8 sub-models)}
    }
    \epsplot{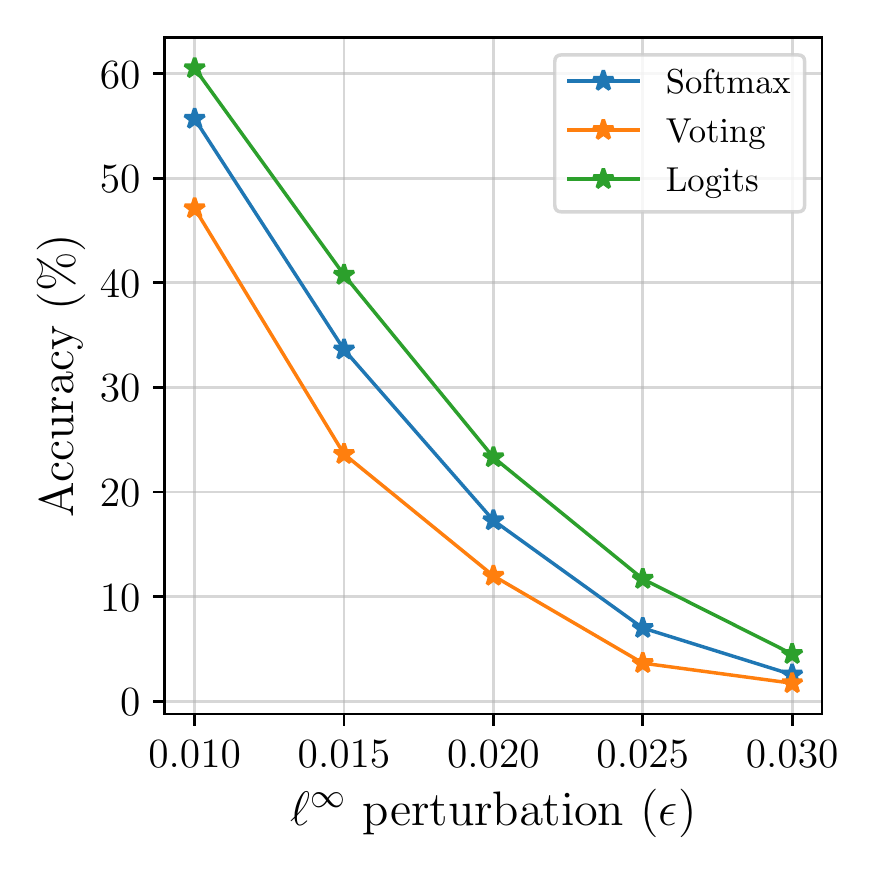}{Diverge}
    \epsplot{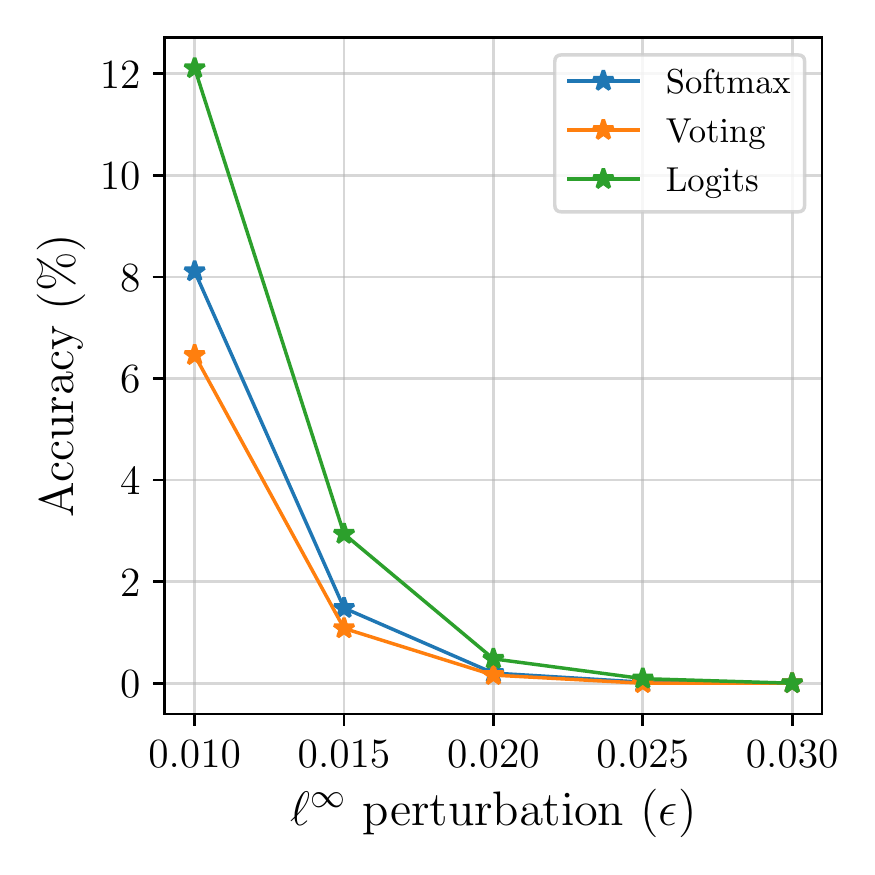}{TRS}
    \epsplot{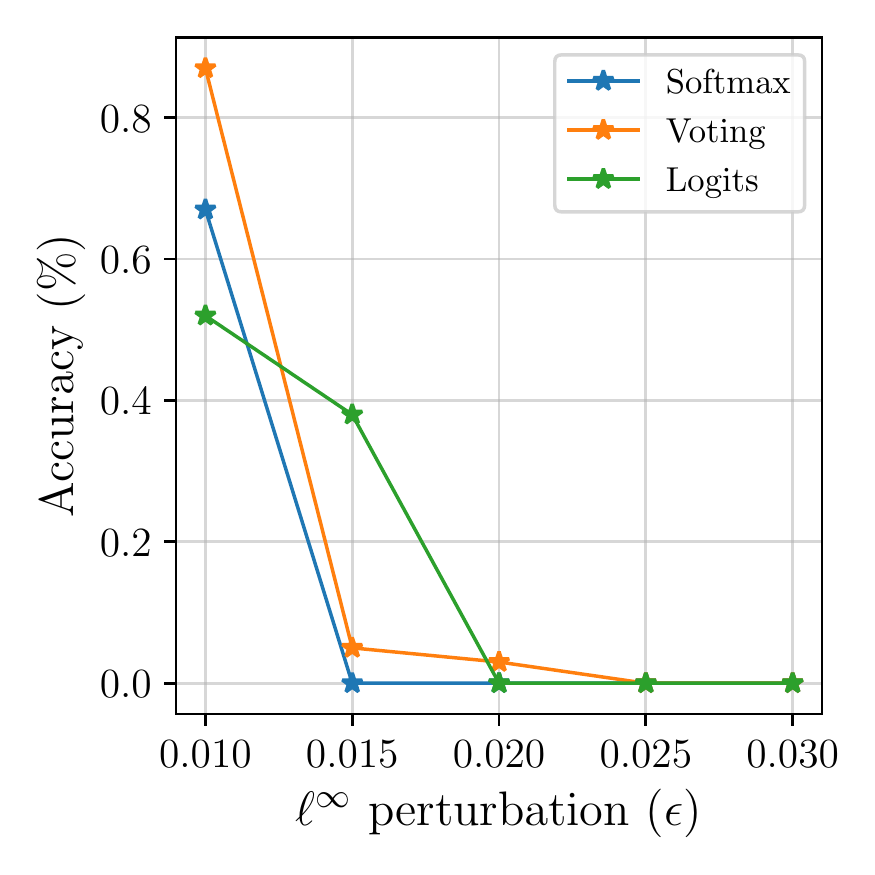}{GAL}
    \caption{%
        Ensemble defenses with 5 sub-models \wrt{}
        the \( \epsilon \) bound on \( \linf \) perturbation.
        Contrary to existing literatures
        that propose ``softmax''-based ensembles~\cite{
            kariyappa2019gal,yang2020dverge,yang2021trs},
        we generally find ``logits''
        to be the most robust option
        across the three ensemble-forming strategies.
        All defenses are evaluated with \attack{} (500 iterations).
        Note that the only exceptions to this rule,
        \ie{}, GAL ensembles with 3 sub-models,
        exhibit very low robust accuracies.
    }\label{fig:epsilon}
\end{figure}

Finally,
\Cref{fig:epsilon}
compares the different ensemble-forming strategies
under an increasing \( \linf \) perturbation bound
\( \epsilon \in \braces{0.01, 0.015, 0.02, 0.025, 0.03} \).
Note that across all configurations,
logit-based ensembles are the most robust.
\input{tables/cifar10_0.02}
\input{tables/cifar10_0.03}%

\subsection{%
    Convergence Speed Comparisons
}\label{app:results:convergence}

\Cref{fig:convergence:dverge,fig:convergence:gal}
compare the convergence speed
of \attack{} against SOTA attacks
for up to 500 iterations.
As AA comprises multiple attack strategies,
we extract its two gradient-based attacks
(APGD-\{CE,DLR\}) to facilitate comparisons.
Each attack uses 5 restarts,
each with up to 100 iterations.
The horizontal axes
show the iteration counts used,
and the vertical axes
denote the percentages of remaining unsuccessful images.
When forming ensembles with logits,
\attack{} and ``No reweighing'' are both identical.
Note that \attack{} substantially
outperforms most existing attacks
under 500 iterations,
and generally requires (up to \(70\times\)) fewer
iterations to achieve the same accuracies
of other attacks with 500 iterations.
\newcommand{\convplot}[3]{%
    \subplot[app:]{convergence/#1_3_#2}{3 sub-models (#3)}
    \subplot[app:]{convergence/#1_5_#2}{5 sub-models (#3)}
    \subplot[app:]{convergence/#1_8_#2}{8 sub-models (#3)}
}%
\begin{figure}[ht]
    \convplot{dverge}{softmax}{softmax}
    \convplot{dverge}{argmax}{voting}
    \convplot{dverge}{logits}{logits}
    \caption{%
        Comparing the convergence speed
        on Dverge~\cite{yang2020dverge} models
        with three ensemble-forming strategies
        (softmax, voting, logits)
        of \attack{}
        against a variant of \attack{}
        without sub-model reweighing,
        \cw, PGD and AA losses (APGD-\{DLR,CE\}).
        For logit-based ensembles
        (\subref{fig:app:convergence/dverge_3_logits},
         \subref{fig:app:convergence/dverge_5_logits},
         \subref{fig:app:convergence/dverge_8_logits}),
        \attack{} and ``No reweighing'' are identical.
        The horizontal and vertical axes
        respectively show the iteration count used,
        and the percentage of unsuccessful images remaining.
        We annotated the number of iterations
        for \attack{} to overtake competition
        with 500 iterations.
    }\label{fig:convergence:dverge}
\end{figure}
\begin{figure}[ht]
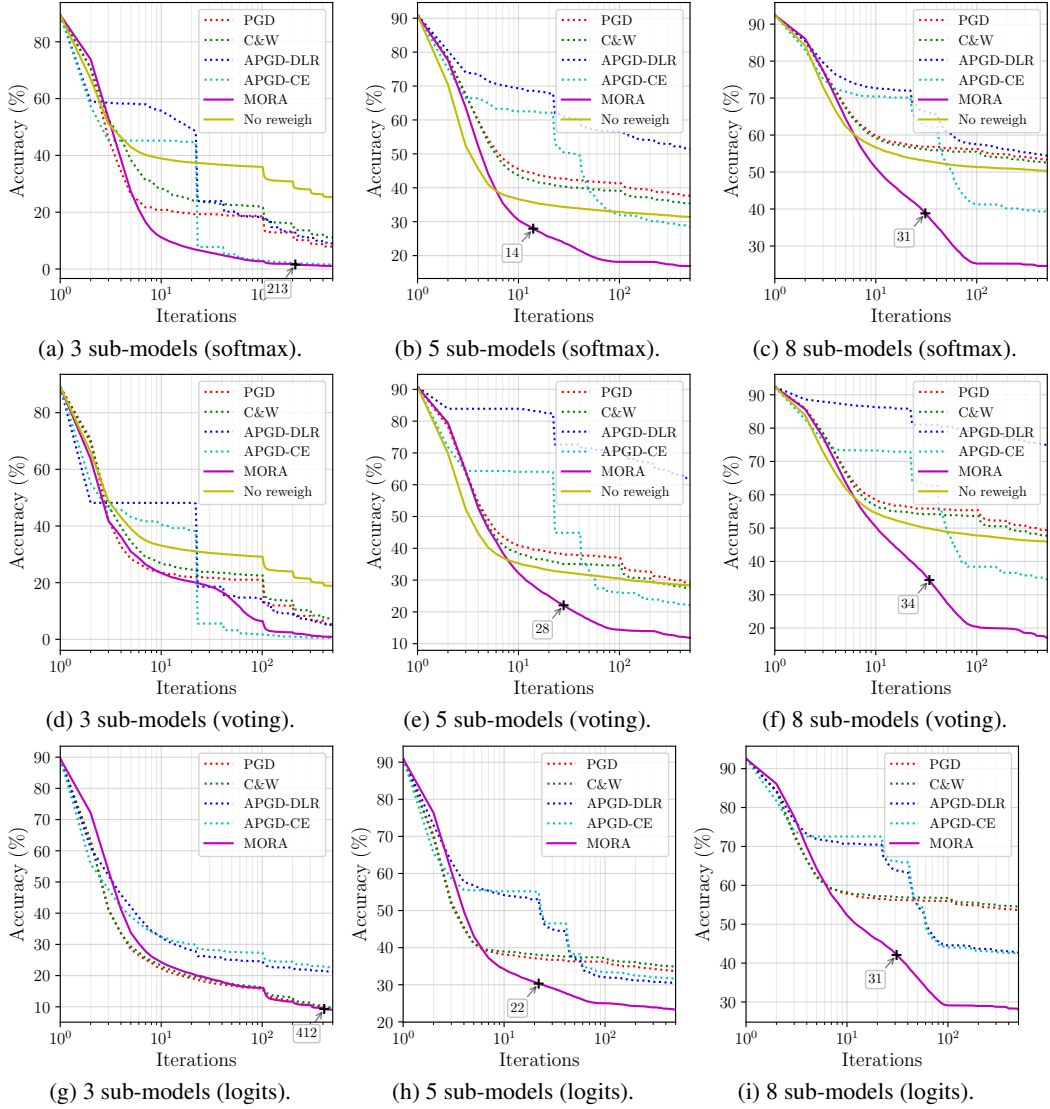

    \convplot{gal}{softmax}{softmax}
    \convplot{gal}{argmax}{voting}
    \convplot{gal}{logits}{logits}
    \caption{%
        Similar comparisons of convergence rates
        on GAL~\cite{kariyappa2019gal} models.
        Please refer to~\Cref{fig:convergence:dverge}
        for a detailed description of the setup.
    }\label{fig:convergence:gal}
\end{figure}

\subsection{%
    Ablation and Sensitivity Analyses
}\label{app:results:ablation}

\input{figures/sensitivity/temperature}
\input{figures/sensitivity/beta}
In \Cref{tab:ablation},
we perform ablation
of the individual components
used in \attack.
We begin with
the standard ``PGD'' attack~\cite{madry18}
with 5 random restarts, each with 100 iterations,
and a constant step size of \( \epsilon / 4 \).
Each row then consecutively
adds a new component.
``Momentum'' introduces momentum \( \momentum = 0.75 \)
as used in~\Cref{alg:overview},
``Cosine Step Size''
then replaces the constant step size
with a cosine schedule \(
    \alpha_i =\epsilon \parens*{
        1+\cos \parens*{ \nicefrac{i\pi}{I} }
    }
\).
``Sub-model Logits'' further
exploits the sub-model logits directly
following~\Cref{sec:method:overview},
and replaces random restarts
with a \( \beta \in \braces{0, 0.25, 0.5, 0.75, 1} \)
grid search
to match the cost of 500 iterations.
``Logit Normalization''
incorporates the normalization
of logits as proposed in~\eqref{eq:scenorm} of~\Cref{sec:method:loss}.
``Adaptive Reweighing''
adaptively adjusts sub-model weights
for attacking ensemble defenses
(\eqref{eq:weight} of~\Cref{sec:method:weight}).
Finally,
``Multiple Targets''~\cite{gowal19surrogate}
additionally uses 100 iterations
of the targeted variant of \( \attackloss \)
on each of the remaining 9 class labels
(\Cref{sec:method:loss}).
\input{tables/ablation}

Increasing the temperature coefficient \( \tau \)
from 1 as used in~\eqref{eq:attackloss}
can affect the convergence speed
of \attack.
To ensure a fair comparison in our results,
we fix a constant \( \tau = 5 \)
for softmax-based ensembles,
and \( \tau = 10 \)
for voting,
as increasing \( \tau \)
may help improve convergence.
In addition,
and \Cref{fig:sensitivity:temperature}
provides sensitivity analyses
of \( \tau \) on the three defending methods
(ADP, GAL, Dverge).

Finally,
\Cref{fig:sensitivity:beta}
shows the effect of varying \( \beta \in [0, 1] \),
the interpolation between \( \attackloss \)
and the ensemble's original loss \( \sceloss \).
Introducing \( \attackloss \)
substantially improves
the strength of attack.
Note that in our comparison results
(\Cref{%
    tab:compare:cifarx,tab:compare:cifarx:at,%
    tab:compare:cifarx:0.02,tab:compare:cifarx:0.03,%
    tab:compare:cifarc}),
instead of 5 random restarts
we use a \(
    \beta \in \braces{0, 0.25, 0.5, 0.75, 1}
\) schedule
to further improve the final attack success rate.

\subsection{\cifarc}\label{app:results:cifarc}

\Cref{tab:compare:cifarc}
compares the attacks
on PDD+DEG~\cite{huang2021pdd} defenses
trained on \cifarc{}.
Similar to GAL~\cite{kariyappa2019gal} and TRS~\cite{yang2021trs},
PDD+DEG also diversifies sub-models gradients
by minimizing cosine-similarities of gradients
via regularization,
and further diversifies feature selection
with adaptive dropouts.
We report attacks on ensembles
with three ensemble-forming methods
(softmax, voting and logits).
Note that we only included PDD+DEG
as other methods did not train on \cifarc.
\input{tables/cifar100}

\subsection{%
    Failure Modes in Ensemble Defenses
}\label{app:results:obfuscation}

\input{figures/loss_surface}
\Cref{fig:loss_surface}
provides a visualization
of the loss surfaces
of ADP~\cite{pang2019adp}
under 3 different ensemble-forming strategies.
In this section,
we continue the discussion
of the two failure modes
in ensemble defense
that induce overestimated robustness
as introduced in~\Cref{sec:intro}:

(a) \emph{%
    Gradient obfuscation
    via ensemble-forming strategies.
}
It is evident that under PGD-10 attacks,
both softmax- and voting-based ensembles
(Figures~\ref{fig:loss:pgd:softmax}
 and~\ref{fig:loss:pgd:voting} respectively)
exhibit to some extent gradient obfuscation
as they result in flatter loss surfaces
in the adversarial direction \( \g \),
whereas the logits-based variant
does not (\Cref{fig:loss:pgd:logits}).

(b) \emph{Gradient diversification.}
As sub-model gradients counteract,
PGD attacks on softmax- and voting-based ensembles
may result in an averaged gradient direction \( \g \)
that experience difficulty in increasing loss
(Figures~\ref{fig:loss:pgd:softmax}
 and~\ref{fig:loss:pgd:voting} respectively).
Adopting sub-model reweighing
(bottom row in~\Cref{fig:loss_surface})
alleviates this difficulty,
and allows the attack to succeed more reliably.

\section{%
    Limitations and Potential Societal Impacts
}\label{app:limitations}

The \( \linf \) white-box threat model
assumes the availability of the models' gradients,
which could present a challenge
as such information may not be available
to the attacker.
It is thus critical
to evaluate white-box robustness
accurately,
as it provides the lower bounds
on the robustness of ensemble defenses
in practical scenarios.

We acknowledge that
adversarial attacks
may have the potential
to be used by a malicious party,
but we believe defending against such attacks
is critically pertinent
to the accurate evaluation of robustness.
We hope this paper furthers
the understanding of ensemble robustness,
and accurately evaluating adversarial robustness
can help improve future defenses.
It is also noteworthy
that the white-box threat model
has applications in the context of advancements
in deep learning,
and can improve
\eg{}, transfer learning~\cite{
    utrera2021adversariallytrained,deng2021adversarial},
GAN training~\cite{bashkirova2019adversarial},
interpretability~\cite{ross2018improving},
and \etc{}

\section{Computational Resources}\label{app:resources}

On NVIDIA Tesla V100 GPUs,
\attack{} with 500 iterations
uses up to \( 1.0 \) GPU-hours
on each ensemble defense,
and \attackmt{} uses up to \( 2.8 \) GPU-hours
on the \cifarx{} test set.
The run time
depends on the number of sub-models
in an ensemble
and its attack difficulty
(\Cref{tab:run_time}).
\begin{table}[ht]
\centering\caption{%
    Run time of \attack{} (up to 500 iterations)
    and \attackmt{} (up to 1.4k iterations)
    for ADP, Dverge and GAL ensemble defenses
    on the \cifarx{} test set.
    Easier ensembles consume less time
    with early stopping.
}\label{tab:run_time}
\begin{tabular}{cc|rr}
    \toprule
    \thead{Run time} (min) & \# & \attack{} & \attackmt{} \\
    \midrule
        & 3 &  2.0 &   3.7 \\
    ADP~\cite{pang2019adp}
        & 5 &  3.4 &   5.8 \\
        & 8 &  6.5 &  11.4 \\
    \midrule
        & 3 & 12.3 &  31.1 \\
    Dverge~\cite{yang2020dverge}
        & 5 & 29.7 &  71.2 \\
        & 8 & 61.9 & 169.3 \\
    \midrule
        & 3 &  2.6 &   4.3 \\
    GAL~\cite{kariyappa2019gal}
        & 5 & 15.4 &  35.2 \\
        & 8 & 37.6 &  91.8 \\
    \bottomrule
\end{tabular}
\end{table}

\section{Licenses}\label{app:licenses}

\Cref{tab:sources}
lists the relevant resources
used in this paper and their respective licenses.
\begin{table}[ht]
\centering\caption{%
    Open-source resources used in this paper.
}\label{tab:sources}
\begin{tabular}{ccl}
    \toprule
    \thead{Name} & \thead{License} & \thead{URL} \\
    \midrule
    PyTorch & BSD
        & \href{https://github.com/pytorch/pytorch}{GitHub: pytorch/pytorch} \\
    Dverge & \tna{}
        & \href{https://github.com/zjysteven/DVERGE}{GitHub: zjysteven/DVERGE} \\
    TRS & \tna{}
        & \href{https://github.com/AI-secure/Transferability-Reduced-Smooth-Ensemble}{GitHub: AI-secure/Transferability-Reduced-Smooth-Ensemble} \\
    \cifarx{} & \tna{}
        & \url{https://www.cs.toronto.edu/~kriz/cifar.html} \\
    \cifarc{} & \tna{}
        & \url{https://www.cs.toronto.edu/~kriz/cifar.html} \\
    \bottomrule
\end{tabular}
\end{table}

%% file: tables/cifar10_0.02.tex
\begin{table}[ht]
\centering%
\caption{%
    Comparing the accuracies
    of SOTA attacks and \attack{}
    on various defenses.
    Please refer to \Cref{tab:compare:cifarx}
    for a detailed explanation.
    This table uses \( \epsilon = 0.02 \)
    as the \( \linf \) perturbation bound,
    under which most defenses (\( 27/48 \))
    have close to nil robustness (\( \leq 2\% \)).
}\label{tab:compare:cifarx:0.02}
\adjustbox{width=0.9\textwidth}{%
\begin{tabular}{cc|cc|ccc|ccc|c}
\toprule
\tbox{\textbf{Defense} \\ Complexity}
    & \textbf{\#}
    & \tbox{\textbf{Clean} \\ 1}
    & \tbox{\textbf{Nominal} \\ \tna}
    & \tbox{\textbf{PGD} \\ 500} & \tbox{\textbf{CW} \\500}
    & \tbox{\textbf{\attack} \\500}
    & \tbox{\textbf{AA} \\4.9k} & \tbox{\textbf{CAA} \\1.8k}
    & \tbox{\textbf{\attackmt} \\1.4k}
    & \( \bm\Delta \) \\
\midrule
\multicolumn{11}{c}{\textbf{Softmax}} \\
\midrule
\multirow{3}{*}{ADP}
    & 3 & 92.88 & 12.53
        & 0.07 & 0.23 & \bfzero{}
        & \bfzero{} & 0.03
        & \bfzero{} & 12.53 \\
    & 5 & 93.34 & 12.75
        & 0.31 & 0.49 & \bfzero{}
        & 0.01 & 0.10
        & \bfzero{} & 12.75 \\
    & 8 & 93.48 & 12.61
        & 3.04 & 1.68 & \bfzero{}
        & 0.02 & 0.47
        & \bfzero{} & 12.61 \\
\midrule
\multirow{3}{*}{Dverge}
    & 3 & 91.99 & 12.78
        & 10.16 & 8.09 & 0.95
        & 6.33 & 6.77
        & \textbf{0.78} & 12.00 \\
    & 5 & 92.38 & 22.36
        & 20.33 & 18.88 & 5.31
        & 10.12 & 16.13
        & \textbf{4.73} & 17.63 \\
    & 8 & 91.65 & 28.20
        & 26.66 & 26.65 & 17.29
        & 22.98 & 25.40
        & \textbf{16.94} & 11.26 \\
\midrule
\multirow{3}{*}{GAL}
    & 3 & 89.41 & 1.75
        & 0.12 & 0.35 & \bfzero{}
        & \bfzero{} & \bfzero{}
        & \bfzero{} & 1.75 \\
    & 5 & 90.93 & 6.80
        & 4.81 & 4.22 & 0.64
        & 2.87 & 2.77
        & \textbf{0.49} & 6.31 \\
    & 8 & 92.45 & 12.22
        & 10.75 & 10.80 & 3.43
        & 8.71 & 6.94
        & \textbf{2.97} & 9.25 \\
\midrule
\multirow{3}{*}{TRS\supdagger}
    & 3 & 70.02 & 2.24
        & 1.49 & 0.80 & 0.20
        & 0.33 & 0.86
        & \textbf{ 0.18} & 2.06 \\
    & 5 & 69.00 & 3.56
        & 1.91 & 1.75 & 0.72
        & 1.07 & 1.68
        & \textbf{0.64} & 2.92 \\
    & 8 & 73.01 & 4.18
        & 1.71 & 1.77 & 1.14
        & 1.50 & 1.72
        & \textbf{0.97} & 3.21 \\
\midrule
\multicolumn{11}{c}{\textbf{Voting}} \\
\midrule
\multirow{3}{*}{ADP}
    & 3 & 91.84 & 22.18\supdagger
        & 0.35 & 0.39 & \bfzero{}
        & 0.19 & 0.41
        & \bfzero{} & 22.18 \\
    & 5 & 93.13 & 21.76\supdagger
        & 0.80 & 0.96 & \bfzero{}
        & 0.58 & 0.67
        & \bfzero{} & 21.76 \\
    & 8 & 93.28 & 15.19\supdagger
        & 3.47 & 2.07 & 0.02
        & 0.90 & 1.15
        & \bfzero{} & 15.19 \\
\midrule
\multirow{3}{*}{Dverge}
    & 3 & 91.72 & 4.70\supdagger
        & 2.61 & 1.54 & 1.11
        & 1.47 & 2.38
        & \textbf{0.89} & 3.81 \\
    & 5 & 92.18 & 9.30\supdagger
        & 7.68 & 6.38 & 5.15
        & 7.15 & 8.33
        & \textbf{4.84} & 4.46 \\
    & 8 & 91.58 & 18.15\supdagger
        & 16.82 & 15.81 & 11.99
        & 15.56 & 17.16
        & \textbf{11.47} & 6.68 \\
\midrule
\multirow{3}{*}{GAL}
    & 3 & 89.09 & 3.57\supdagger
        & 0.23 & 0.11 & \bfzero{}
        & \bfzero{} & \bfzero{}
        & \bfzero{} & 3.57 \\
    & 5 & 90.77 & 3.26\supdagger
        & 1.95 & 1.83 & 0.54
        & 1.64 & 2.10
        & \textbf{0.48} & 2.78 \\
    & 8 & 92.37 & 9.36\supdagger
        & 7.26 & 6.85 & 2.34
        & 4.91 & 5.33
        & \textbf{1.90} & 7.46 \\
\midrule
\multirow{3}{*}{TRS\supdagger}
    & 3 & 68.95 & 0.83
        & 0.53 & 0.25 & 0.16
        & 0.59 & 0.34
        & \textbf{ 0.12} & 0.71 \\
    & 5 & 68.31 & 1.34
        & 0.92 & 0.83 & 0.60
        & 0.72 & 0.91
        & \textbf{ 0.53} & 0.81 \\
    & 8 & 72.05 & 1.45
        & 1.17 & 0.87 & 0.86
        & 0.71 & 1.00
        & \textbf{0.68} & 0.77 \\
\midrule
\multicolumn{11}{c}{\textbf{Logits}} \\
\midrule
\multirow{3}{*}{ADP}
    & 3 & 92.86 & 0.13\supdagger{}
        & \bfzero{} & \bfzero{} & \bfzero{}
        & \bfzero{} & \bfzero{}
        & \bfzero{} & 0.13 \\
    & 5 & 93.48 & 0.20\supdagger{}
        & \bfzero{} & 0.04 & \bfzero{}
        & \bfzero{} & \bfzero{}
        & \bfzero{} & 0.20 \\
    & 8 & 93.38 & 0.04\supdagger{}
        & \bfzero{} & \bfzero{} & \bfzero{}
        & \bfzero{} & \bfzero{}
        & \bfzero{} & 0.04 \\
\midrule
\multirow{3}{*}{Dverge}
    & 3 & 92.19 & 4.39\supdagger{}
        & 4.12 & 4.24 & 3.27
        & 3.31 & 3.40
        & \textbf{3.17} & 1.22 \\
    & 5 & 92.28 & 12.66\supdagger{}
        & 12.24 & 12.56 & 10.83
        & 10.93 & 10.99
        & \textbf{10.80} & 1.86 \\
    & 8 & 91.73 & 24.75\supdagger{}
        & 24.41 & 25.42 & 23.32
        & 23.39 & 23.49
        & \textbf{23.31} & 1.54 \\
\midrule
\multirow{3}{*}{GAL}
    & 3 & 89.50 & 1.10\supdagger{}
        & 0.25 & 0.22 & \bfzero{}
        & \bfzero{} & \bfzero{}
        & \bfzero{} & 1.10 \\
    & 5 & 90.93 & 3.98\supdagger{}
        & 3.14 & 3.93 & 1.01
        & 0.47 & 0.64
        & \textbf{0.45} & 3.53 \\
    & 8 & 92.54 & 11.18\supdagger{}
        & 9.81 & 10.81 & 3.86
        & 3.52 & 3.79
        & \textbf{3.41} & 7.77 \\
\midrule
\multirow{3}{*}{TRS\supdagger{}}
    & 3 & 69.72 & 0.73
        & 0.64 & 0.71 & 0.48
        & 0.50 & 0.52
        & \textbf{0.47} & 0.26 \\
    & 5 & 68.90 & 1.60
        & 1.51 & 1.63 & 1.35
        & 1.31 & 1.33
        & \textbf{1.30} & 0.30 \\
    & 8 & 72.24 & 1.78
        & 1.69 & 1.78 & 1.52
        & 1.47 & 1.46
        & \textbf{1.42} & 0.36 \\
\bottomrule
\end{tabular}\vspace{-20pt}}
\end{table}%

%% file: tables/cifar10_0.03.tex
\begin{table}[!ht]
\centering%
\caption{%
    Comparing the accuracies
    of SOTA attacks and \attack{}
    on various defenses.
    Please refer to \Cref{tab:compare:cifarx}
    for a detailed explanation.
    This table uses \( \epsilon = 0.03 \)
    as the \( \linf \) perturbation bound,
    under which almost all defenses exhibit \( 0\% \) robustness,
    and all defenses fail to give \( 5\% \) robustness.
}\label{tab:compare:cifarx:0.03}
\adjustbox{width=0.9\textwidth}{%
\begin{tabular}{cc|cc|ccc|ccc|c}
\toprule
\tbox{\textbf{Defense} \\ Complexity}
    & \textbf{\#}
    & \tbox{\textbf{Clean} \\ 1}
    & \tbox{\textbf{Nominal} \\ \tna}
    & \tbox{\textbf{PGD} \\ 500} & \tbox{\textbf{CW} \\500}
    & \tbox{\textbf{\attack} \\500}
    & \tbox{\textbf{AA} \\4.9k} & \tbox{\textbf{CAA} \\1.8k}
    & \tbox{\textbf{\attackmt} \\1.4k}
    & \( \bm\Delta \) \\
\midrule
\multicolumn{11}{c}{\textbf{Softmax}} \\
\midrule
\multirow{3}{*}{ADP}
    & 3 & 92.88 & 7.29
        & \bfzero{} & \bfzero{} & \bfzero{}
        & \bfzero{} & \bfzero{}
        & \bfzero{} & 7.29 \\
    & 5 & 93.34 & 8.52
        & 0.04 & 0.03 & \bfzero{}
        & \bfzero{} & \bfzero{}
        & \bfzero{} & 8.52 \\
    & 8 & 93.48 & 9.64
        & 1.75 & 0.89 & \bfzero{}
        & \bfzero{} & 0.05
        & \bfzero{} & 9.64 \\
\midrule
\multirow{3}{*}{Dverge}
    & 3 & 91.99 & 2.91
        & 1.43 & 1.47 & 0.02
        & 1.59 & 1.10
        & \bfzero{} & 2.91 \\
    & 5 & 92.38 & 5.90
        & 4.69 & 4.27 & 1.13
        & 2.98 & 4.05
        & \textbf{0.23} & 5.67 \\
    & 8 & 91.65 & 9.17
        & 8.28 & 8.46 & 2.54
        & 6.84 & 8.21
        & \textbf{2.34} & 6.83 \\
\midrule
\multirow{3}{*}{GAL}
    & 3 & 89.41 & 1.26
        & 0.01 & 0.06 & \bfzero{}
        & \bfzero{} & \bfzero{}
        & \bfzero{} & 1.26 \\
    & 5 & 90.93 & 0.68
        & 0.34 & 0.22 & \bfzero{}
        & 0.20 & 0.10
        & \bfzero{} & 0.68 \\
    & 8 & 92.45 & 1.18
        & 0.92 & 0.88 & 0.27
        & 1.19 & 0.79
        & \textbf{0.17} & 1.01 \\
\midrule
\multirow{3}{*}{TRS\supdagger}
    & 3 & 70.02 & 0.71
        & 0.11 & 0.04 & \bfzero{}
        & 0.01 & 0.06
        & \bfzero{} & 0.71 \\
    & 5 & 69.00 & 2.24
        & 0.18 & 0.17 & 0.04
        & \textbf{0.03} & 0.14
        & \textbf{0.03} & 2.21 \\
    & 8 & 73.01 & 2.32
        & 0.12 & 0.17 & 0.06
        & 0.09 & 0.15
        & \textbf{0.02} & 2.30 \\
\midrule
\multicolumn{11}{c}{\textbf{Voting}} \\
\midrule
\multirow{3}{*}{ADP}
    & 3 & 91.84 & 15.18\supdagger
        & 0.03 & 0.09 & \bfzero{}
        & 0.02 & 0.03
        & \bfzero{} & 15.18 \\
    & 5 & 93.13 & 14.97\supdagger
        & 0.33 & 0.17 & \bfzero{}
        & 0.03 & 0.04
        & \bfzero{} & 14.97 \\
    & 8 & 93.28 & 11.26\supdagger
        & 1.91 & 1.01 & \bfzero{}
        & 0.09 & 0.17
        & \bfzero{} & 11.26 \\
\midrule
\multirow{3}{*}{Dverge}
    & 3 & 91.72 & 0.99\supdagger
        & 0.19 & 0.07 &0.06
        & 0.05 & 0.10
        & \bfzero{} & 0.99 \\
    & 5 & 92.18 & 0.98\supdagger
        & 0.55 & 0.44 & 0.33
        & 0.50 & 0.71
        & \textbf{0.23} & 0.75 \\
    & 8 & 91.58 & 3.42\supdagger
        & 2.72 & 2.27 & 1.72
        & 2.30 & 2.92
        & \textbf{1.56} & 1.86 \\
\midrule
\multirow{3}{*}{GAL}
    & 3 & 89.09 & 1.90\supdagger
        & 0.02 & 0.04 & \bfzero{}
        & \bfzero{} & \bfzero{}
        & \bfzero{} & 1.90 \\
    & 5 & 90.77 & 0.27\supdagger
        & 0.10 & 0.07 & \bfzero{}
        & 0.04 & 0.07
        & \bfzero{} & 0.27 \\
    & 8 & 92.37 & 0.66\supdagger
        & 0.44 & 0.38 & 0.15
        & 0.36 & 0.45
        & \textbf{0.13} & 0.53 \\
\midrule
\multirow{3}{*}{TRS\supdagger}
    & 3 & 68.95 & 0.03
        & 0.02 & \bfzero{} & \bfzero{}
        & \bfzero{} & \bfzero{}
        & \bfzero{} & 0.03 \\
    & 5 & 68.31 & 0.16
        & 0.10 & 0.04 & 0.06
        & 0.01 & 0.03
        & \bfzero{} & 0.09 \\
    & 8 & 72.05 & 0.15
        & \bfzero{} & 0.01 & \bfzero{}
        & \bfzero{} & \bfzero{}
        & \bfzero{} & 0.15 \\
\midrule
\multicolumn{11}{c}{\textbf{Logits}} \\
\midrule
\multirow{3}{*}{ADP}
    & 3 & 92.86 & 0.01\supdagger
        & \bfzero{} & \bfzero{} & \bfzero{}
        & \bfzero{} & \bfzero{}
        & \bfzero{} & 0.01 \\
    & 5 & 93.48 & 0.05\supdagger
        & \bfzero{} & \bfzero{} & \bfzero{}
        & \bfzero{} & \bfzero{}
        & \bfzero{} & 0.05 \\
    & 8 & 93.38 & \bfzero\supdagger
        & \bfzero{} & \bfzero{} & \bfzero{}
        & \bfzero{} & \bfzero{}
        & \bfzero{} & 0.00 \\
\midrule
\multirow{3}{*}{Dverge}
    & 3 & 92.19 & 0.18\supdagger
        & 0.15 & 0.15 & \textbf{0.09}
        & \textbf{0.09} & \textbf{0.09}
        & \textbf{0.09} & 0.09 \\
    & 5 & 92.28 & 1.25\supdagger
        & 1.13 & 1.13 & 0.83
        & \textbf{0.79} & 0.83
        & \textbf{0.79} & 0.46 \\
    & 8 & 91.73 & 6.01\supdagger
        & 5.77 & 5.95 & 4.51
        & 4.54 & 4.63
        & \textbf{4.42} & 1.59 \\
\midrule
\multirow{3}{*}{GAL}
    & 3 & 89.50 & 0.09\supdagger
        & \bfzero{} & 0.02 & \bfzero{}
        & \bfzero{} & \bfzero{}
        & \bfzero{} & 0.09 \\
    & 5 & 90.93 & 0.24\supdagger
        & 0.16 & 0.13 & \bfzero{}
        & \bfzero{} & \bfzero{}
        & \bfzero{} & 0.24 \\
    & 8 & 92.54 & 0.66\supdagger
        & 0.55 & 0.70 & 0.22
        & \textbf{0.17} & 0.18
        & \textbf{0.17} & 0.48 \\
\midrule
\multirow{3}{*}{TRS\supdagger}
    & 3 & 69.72 & 0.01
        &0.01 & 0.02 & \bfzero{}
        & 0.01 & 0.01
        & \bfzero{} & 0.01 \\
    & 5 & 68.90 & 0.11
        & 0.10 & 0.12 & 0.09
        & 0.07 & 0.06
        & \textbf{0.04} & 0.07 \\
    & 8 & 72.24 & 0.13
        & 0.12 & 0.15 & 0.07
        & 0.08 & 0.07
        & \textbf{0.06} & 0.07 \\
\bottomrule
\end{tabular}
}\end{table}

%% file: figures/sensitivity/temperature.tex
\begin{figure}[ht]
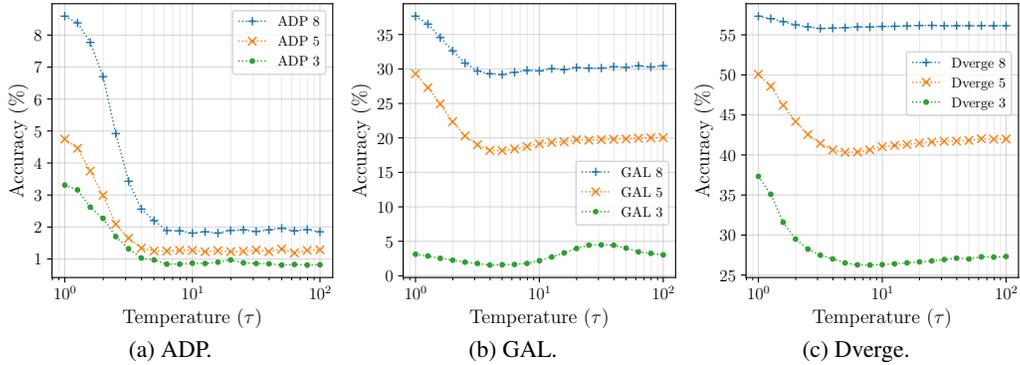

    \subplot{sensitivity/temperature/adp}{ADP}
    \subplot{sensitivity/temperature/gal}{GAL}
    \subplot{sensitivity/temperature/dverge}{Dverge}
    \caption{%
        Sensitivity analyses
        of the temperature coefficient \( \tau \)
        as introduced in~\eqref{eq:weight}
        and used in \( \attackloss \).
        All ensembles use softmax values
        of sub-models to form outputs,
        as this is used by
        ADP~\cite{pang2019adp}, GAL~\cite{kariyappa2019gal}
        and Dverge~\cite{yang2020dverge}.
        All defending ensembles share similar flat regions
        of optimal \( \tau \) with low sensitivity.
    }\label{fig:sensitivity:temperature}
\end{figure}

%% file: figures/sensitivity/beta.tex
\begin{figure}[ht]
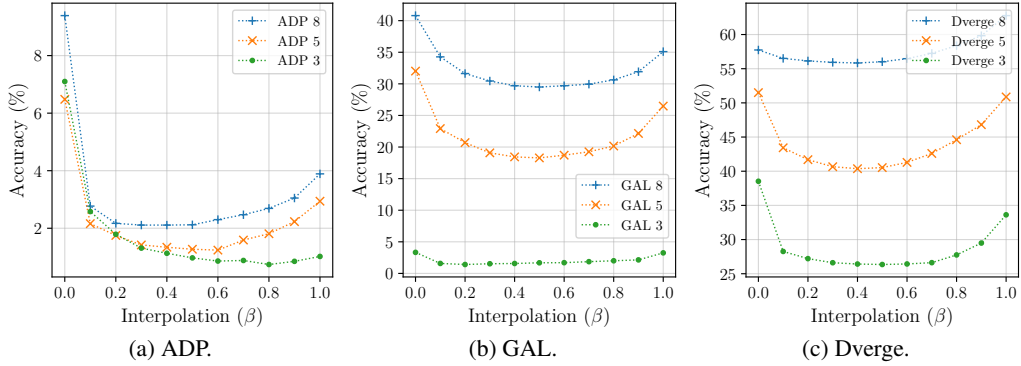

    \subplot{sensitivity/beta/adp}{ADP}
    \subplot{sensitivity/beta/gal}{GAL}
    \subplot{sensitivity/beta/dverge}{Dverge}
    \caption{%
        Sensitivity analyses
        of the \( \beta \) interpolation
        between the original ensemble loss \( \sceloss \)
        and \( \attackloss \),
        as introduced in~\eqref{eq:attackloss}.
        All ensembles use softmax values
        of sub-models to form outputs,
        as this is used by
        ADP~\cite{pang2019adp}, GAL~\cite{kariyappa2019gal}
        and Dverge~\cite{yang2020dverge}.
        All defending ensembles share similar flat regions
        of optimal \( \beta \) with low sensitivity.
    }\label{fig:sensitivity:beta}
\end{figure}

%% file: tables/ablation.tex
\begin{table*}[bt]
\centering%
\caption{%
    Ablation study
    of individual components used
    by \attack.
    ``PGD'' is the standard
    PGD attack~\cite{madry18}
    with 5 restarts
    each with 100 iterations.
    New components are added
    in consecutive rows.
    All rows use 500 iterations at most per image,
    except ``Multiple Targets''
    which additionally
    uses 100 iterations each
    for the remaining 9 class labels.
}\label{tab:ablation}
\adjustbox{width=\textwidth}{%
\begin{tabular}{l|ccc|ccc|ccc|ccc}
\toprule
\multicolumn{1}{c|}{\textbf{Defenses}}
    & \multicolumn{3}{c|}{\textbf{ADP}~\cite{pang2019adp}}
    & \multicolumn{3}{c|}{\textbf{Dverge}~\cite{yang2020dverge}}
    & \multicolumn{3}{c|}{\textbf{GAL}~\cite{kariyappa2019gal}}
    & \multicolumn{3}{c}{\textbf{TRS}~\cite{yang2021trs}} \\
\multicolumn{1}{c|}{\textbf{\#}}
    & 3 & 5 & 8 & 3 & 5 & 8 & 3 & 5 & 8 & 3 & 5 & 8 \\
\midrule
\multicolumn{12}{c}{\textbf{Softmax}} \\
\midrule
{PGD}
    & 8.98 & 9.07 & 10.85
    & 49.59 & 58.13 & 61.31
    & 10.44 & 43.16 & 54.60
    & 15.72 & 17.92 & 19.62 \\
{+ Momentum}
    & 5.98 & 7.10 & 9.22
    & 44.49 & 54.61 & 59.13
    & 8.13 & 37.59 & 53.39
    & 14.01 & 15.91 & 18.02 \\
{+ Early Stop}
    & 5.04 & 6.57 & 8.89
    & 40.18 & 53.53 & 58.84
    & 3.70 & 35.34 & 50.53
    & 13.21 & 15.63 & 17.97 \\
{+ Cosine Step Size}
    & 4.14 & 5.08 & 7.65
    & 38.09 & 50.71 & 58.36
    & 1.34 & 28.60 & 37.17
    & 12.26 & 14.88 & 17.32 \\
{+ Sub-model Logits}
    & 0.69 & 1.13 & 1.84
    & 28.51 & 42.91 & 56.95
    & 1.78 & 20.73 & 30.51
    & 9.15 &  13.52 & 16.80 \\
{+ Logit Normalization}
    & 0.66 & 0.91 & 1.56
    & 27.42 & 41.02 & 55.78
    & 2.39 & 19.13 & 29.71
    & 9.21 & 13.49 & 16.78 \\
{+ Adaptive Reweighing}
    & 0.63 & 0.96 & 1.72
    & 25.87 & 39.98 & 55.61
    & 0.64 & 17.07 & 28.50
    & 8.24 & 12.67 & 15.93 \\
{+ Multiple Targets}
    & \textbf{0.34} & \textbf{0.67} & \textbf{1.32}
    & \textbf{25.26} & \textbf{39.50} & \textbf{55.57}
    & \textbf{0.51} & \textbf{16.05} & \textbf{27.44}
    & \textbf{7.60} & \textbf{12.47} & \textbf{15.64} \\
\midrule
\multicolumn{12}{c}{\textbf{Voting}} \\
\midrule
{PGD}
    & 13.31 & 13.87 & 13.72
    & 36.06 & 49.59 & 57.32
    & 7.23 & 34.71 & 51.37
    & 13.38 & 15.56 & 17.06 \\
{+ Momentum}
    & 9.32 & 12.42 & 12.53
    & 31.48 & 44.28 & 53.72
    & 5.85 & 29.33 & 49.56
    & 10.19 & 12.71 & 14.57 \\
{+ Early Stop}
    & 5.04 & 8.19 & 10.59
    & 28.45 & 41.14 & 51.65
    & 1.38 & 23.68 & 44.92
    & 8.24 & 11.44 & 13.53 \\
{+ Cosine Step Size}
    & 4.54 & 7.55 & 9.17
    & 28.14 & 40.85 & 51.43
    & 0.55 & 19.21 & 29.36
    & 8.07 & 10.94 & 12.87 \\
{+ Sub-model Logits}
    & 0.56 & 1.25 & 2.01
    & 27.98 & 41.32 & 51.78
    & 1.21 & 19.41 & 27.90
    & 7.67 &  11.14 & 13.25 \\
{+ Logit Normalization}
    & 0.73 & 1.15 & 1.81
    & 29.79 & 43.87 & 54.30
    & 3.82 & 21.08 & 29.27
    & 7.68 & 11.13 & 13.29 \\
{+ Adaptive Reweighing}
    & 0.59 & 1.11 & 2.51
    & 23.47 & 34.94 & 47.24
    & 0.85 & 12.88 & 21.62
    & 5.66 & 8.82 & 11.36 \\
{+ Multiple Targets}
    & \textbf{0.29} & \textbf{0.62} & \textbf{1.65}
    & \textbf{22.91} & \textbf{34.46} & \textbf{46.10}
    & \textbf{0.35} & \textbf{12.25} & \textbf{20.16}
    & \textbf{5.44} & \textbf{8.38} & \textbf{10.69} \\
\midrule
\multicolumn{12}{c}{\textbf{Logits}} \\
\midrule
{PGD}
    & 1.55 & 3.29 & 5.32
    & 39.20 & 51.28 & 61.22
    & 4.39 & 38.27 & 54.64
    & 14.17 & 17.78 & 20.39 \\
{+ Momentum}
    & 0.87 & 1.97 & 3.57
    & 39.20 & 50.57 & 60.95
    & 10.01 & 33.97 & 53.67
    & 13.06 & 16.65 & 19.20 \\
{+ Early Stop}
    & 0.82 & 1.77 & 3.27
    & 37.91 & 50.53 & 60.89
    & 8.12 & 31.24 & 50.68
    & 12.96 & 16.51 & 19.13 \\
{+ Cosine Step Size}
    & 0.48 & 1.19 & 2.19
    & 37.22 & 49.93 & 60.66
    & 0.85 & 22.80 & 31.37
    & 12.36 & 16.08 & 18.44 \\
{+ Sub-model Logits}
    & 0.52 & 1.21 & 2.24
    & 37.22 & 49.95 & 60.65
    & 0.81 & 22.79 & 31.25
    & 12.36 &  16.05 & 18.48 \\
{+ Logit Normalization}
    & 0.45 & 1.15 & 2.12
    & 36.90 & 49.63 & 60.53
    & 0.35 & 22.21 & 31.28
    & 12.35 & 16.09 & 18.48 \\
{+ Adaptive Reweighing}
    & 0.47 & 1.14 & 2.07
    & 36.88 & 49.65 & 60.52
    & 0.52 & 22.12 & 31.57
    & 12.10 & 15.85 & 18.22 \\
{+ Multiple Targets}
    & \textbf{0.21} & \textbf{0.89} & \textbf{1.93}
    & \textbf{36.84} & \textbf{49.59} & \textbf{60.49}
    & \textbf{0.03} & \textbf{19.40} & \textbf{30.66}
    & \textbf{12.07} & \textbf{15.82} & \textbf{18.17} \\
\bottomrule
\end{tabular}\vspace{-20pt}}
\end{table*}

%% file: tables/cifar100.tex
\begin{table}[htbp]
\centering
\caption{%
    Comparing the accuracies
    of SOTA attacks and \attack{}
    on PDD+DEG~\cite{huang2021pdd} defenses
    trained with \cifarc.
    Please refer to \Cref{tab:compare:cifarx}
    for a detailed explanation.
    This table uses \( \epsilon = 0.01 \)
    as the \( \linf \) perturbation bound,
    and evaluates its results
    on the \cifarc{} test dataset.
}\label{tab:compare:cifarc}
\adjustbox{width=\textwidth}{%
\begin{tabular}{cc|cc|ccc|cccc|c}
\toprule
\tbox{\textbf{PDD+DEG}~\cite{huang2021pdd}}
    & \textbf{\#}
    & \tbox{\textbf{Clean} \\ 1}
    & \tbox{\textbf{Nominal} \\ \tna}
    & \tbox{\textbf{PGD} \\500} & \tbox{\textbf{CW} \\500}
    & \tbox{\textbf{\attack} \\500}
    & \tbox{\textbf{\aaa} \\12k}
    & \tbox{\textbf{AA} \\4.9k} & \tbox{\textbf{CAA} \\1.8k}
    & \tbox{\textbf{\attackmt} \\1.4k}
    & \( \bm\Delta \) \\
\midrule
Softmax
    & 3 & 79.30 & 22.77
        & 11.32 & 14.19 &  1.33
        &  9.44 &  9.19 & 11.46
        & \textbf{ 0.62} & 22.15 \\
Voting
    & 3 & 79.28 &  6.55\supdagger
        &  2.65 &  2.38 &  0.91
        &  0.66 &  1.16 &  1.29
        & \textbf{ 0.55} &  6.00 \\
Logits
    & 3 & 78.94 &  7.25\supdagger
        &  3.29 &  1.30 &  0.90
        &  0.63 &  0.63 &  0.67
        & \textbf{ 0.56} &  6.69 \\
\bottomrule
\end{tabular}\vspace{-20pt}}
\end{table}

%% file: figures/loss_surface.tex
\begin{figure}[ht]
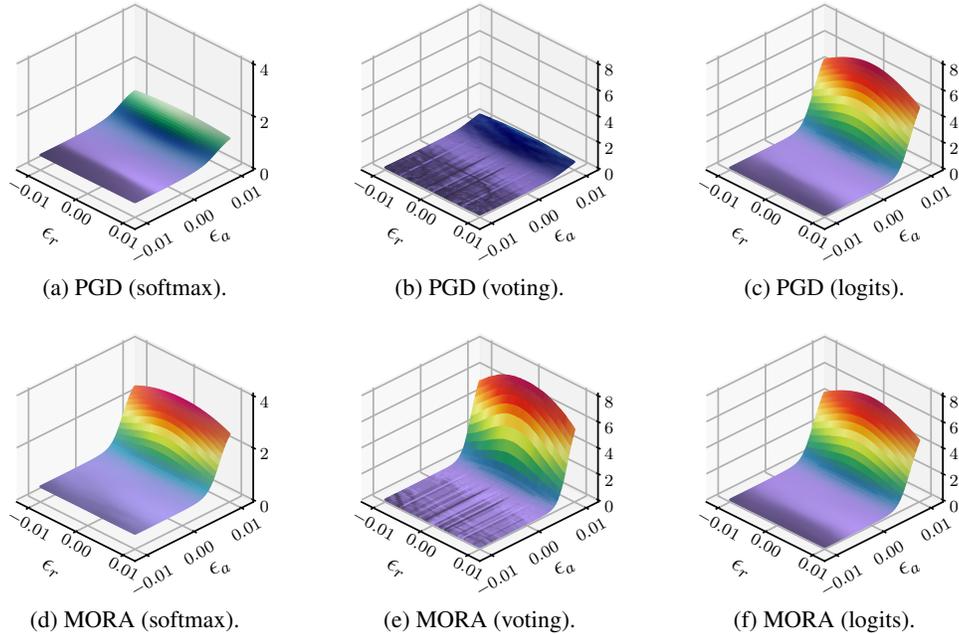

    \centering%
    \newcommand{\lspgdsubfig}[2]{%
        \begin{subfigure}[b]{0.3\textwidth}
            \centering\includegraphics[
                scale=0.8, trim=0 10pt 0 15pt
            ]{loss_surface/#1_adv}%
            \caption{PGD (#2).}\label{fig:loss:pgd:#2}
        \end{subfigure}%
    }
    \newcommand{\lsmorasubfig}[2]{%
        \begin{subfigure}[b]{0.3\textwidth}
            \centering\includegraphics[
                scale=0.8, trim=0 10pt 0 0
            ]{loss_surface/#1_adv_sub_original}%
            \caption{\attack{} (#2).}\label{fig:loss:mora:#2}
        \end{subfigure}%
    }
    \hspace*{\fill}
    \lspgdsubfig{softmax}{softmax}
    \hspace*{\fill}
    \lspgdsubfig{argmax}{voting}
    \hspace*{\fill}
    \lspgdsubfig{logits}{logits}
    \hspace*{\fill}
    \\
    \hspace*{\fill}
    \lsmorasubfig{softmax}{softmax}
    \hspace*{\fill}
    \lsmorasubfig{argmax}{voting}
    \hspace*{\fill}
    \lsmorasubfig{logits}{logits}
    \hspace*{\fill}
    \caption{%
        The averaged loss surfaces
        across all samples
        that resisted PGD-10 attacks
        under all ensemble-forming modes
        of a 3 sub-model ensemble
        trained with ADP~\cite{pang2019adp}
        in the image space
        \(
            \x + \g \epsilon_\mathrm{a}
            +\g^{\bot} \epsilon_\mathrm{r}
        \).
        Here,
        \( \g \) denotes the normalized adversarial direction
        after accumulating 10 initial iterations
        of gradient updates
        at the natural input \( \x \),
        and \( \g^\bot \) is its uniformly randomized orthogonal.
        The top row uses standard PGD attacks,
        and the bottom row then
        replaces the SCE loss used in the top row
        with the \attack{} loss
        and uses \( \beta = 0.5 \).
    }\label{fig:loss_surface}
\end{figure}%